\pdfoutput=1
\documentclass{article}
\usepackage[nonatbib,preprint]{neurips_2021}

\usepackage[utf8]{inputenc} 
\usepackage[T1]{fontenc}    
\usepackage{url}            
\usepackage{amsfonts}       
\usepackage{microtype}      
\usepackage{times}
\usepackage{epsfig}
\usepackage{graphicx}
\usepackage{amsmath,amssymb}
\usepackage{bm} %

\usepackage{algorithm}
\usepackage{listings}
\usepackage[table]{xcolor}
\usepackage{booktabs} %
\usepackage{colortbl}
\usepackage{multirow}
\usepackage[font=small]{caption}

\usepackage{enumitem}

\usepackage{pifont}
\usepackage{xspace}
\usepackage{multirow}

\usepackage[pagebackref=true,breaklinks=true,letterpaper=true,colorlinks,bookmarks=false]{hyperref}

\newif\ifarxiv

\definecolor{myblue}{RGB}{8,48,107}
\definecolor{myred}{RGB}{127,39,4}
\newcommand{\tblue}[1]{{\color{myblue}#1}}

\def \etal {\textit{et al.}\xspace}
\def \eg {\textit{e.g.}\xspace}
\def \ie {\textit{i.e.}\xspace}

\def \vs {\textit{vs.}\xspace}
\def \wrt {{w.r.t.}\xspace}

\def \pzo {\phantom{0}} 
\def \dzo {\phantom{00}} 
\def \tzo {\phantom{000}}

\def \OURS {ResMLP\xspace}

\def \ImNet {ImageNet\xspace}

\author{
\begin{minipage}{\linewidth}
\begin{center}
Hugo Touvron$^{1,2}$ \hspace{0.35cm} Piotr Bojanowski$^{1}$ \hspace{0.35cm} Mathilde Caron$^{1,3}$ \hspace{0.25cm}  \\[0.2cm] 
Matthieu Cord$^{2,4}$ \hspace{0.35cm}  Alaaeldin El-Nouby$^{1,3}$ \hspace{0.35cm} Edouard Grave$^{1}$ \hspace{0.35cm} Gautier Izacard$^{1}$  
\\[0.2cm] 
Armand Joulin$^1$  \hspace{0.35cm}  Gabriel Synnaeve$^{1}$ \hspace{0.35cm}  Jakob Verbeek$^{1}$ \hspace{0.35cm} Herv\'e J\'egou$^{1}$ \\[0.5cm]
\scalebox{1.}{\normalsize \textmd{$^1$Facebook AI\hspace{0.6cm} $^2$Sorbonne University\hspace{0.6cm}  $^3$Inria \hspace{0.6cm}$^4$valeo.ai}}
\\[1cm]
\end{center}
\end{minipage}
}
\title{ResMLP: Feedforward networks for image classification with data-efficient training}

\makeatletter
\let\inserttitle\@title
\makeatother

\makeatletter
\renewcommand{\paragraph}{%
  \@startsection{paragraph}{4}%
  {\z@}{2.8ex \@plus 1ex \@minus .2ex}{-1em}%
  {\normalfont\normalsize\bfseries}%
}
\makeatother

\date{~}
\begin{document}

\maketitle

\begin{abstract}
We  present  \OURS, an architecture built entirely upon multi-layer perceptrons  for image classification. It is a simple residual network that alternates (i) a linear layer in which image patches interact, independently and identically across channels, and (ii)  a two-layer feed-forward network in which channels interact independently per patch. 
When trained with a modern training strategy using heavy data-augmentation and optionally distillation, it attains surprisingly good accuracy/complexity trade-offs on ImageNet. 
We also train \OURS models in a self-supervised setup, to further remove priors from employing a labelled dataset. 
Finally, by adapting our model to machine translation we achieve  surprisingly good results. 

We share pre-trained models and our code based on the Timm library. 
\end{abstract}

\section{Introduction}
\label{sec:introduction}

Recently, the transformer architecture~\cite{vaswani2017attention}, adapted from its original use in natural language processing with only minor changes, has achieved  performance competitive with the state of the art on \ImNet-1k~\cite{Russakovsky2015ImageNet12} when pre-trained with a sufficiently large amount of data~\cite{dosovitskiy2020image}. 
Retrospectively, this achievement is another step towards learning visual features with less priors: 
Convolutional Neural Networks~(CNN) had replaced the hand-designed choices from hard-wired features with flexible and trainable architectures.
Vision transformers further removes several hard decisions encoded in the convolutional architectures, namely the translation invariance and local connectivity.
This evolution toward less hard-coded prior in the architecture has been fueled by better training schemes~\cite{dosovitskiy2020image,Touvron2020TrainingDI}, and, in this paper, we push this trend further by showing that a purely multi-layer perceptron (MLP) based architecture, called Residual Multi-Layer Perceptrons (\OURS), is competitive on image classification.  
\OURS is designed to be simple and encoding little prior about images: it takes image patches as input, projects them with a linear layer, and sequentially updates their representations with two residual operations: (i) a \emph{cross-patch} linear layer applied to all channels independently; and (ii) an \emph{cross-channel} single-layer MLP  applied independently to all patches. 
At the end of the network, the patch representations are average pooled, and fed to a linear classifier.
We outline \OURS in Figure~\ref{fig:architecture} and detail it further in Section~\ref{sec:method}.

The \OURS architecture is strongly inspired by the vision transformers (ViT)~\cite{dosovitskiy2020image}, yet it is much simpler in several ways: we replace the self-attention sublayer by a linear layer, resulting in an architecture with only linear layers and GELU non-linearity~\cite{Hendrycks2016GaussianEL}. 
We observe that the training of \OURS is more stable than ViTs when using the same training scheme as in DeiT~\cite{Touvron2020TrainingDI} and CaiT~\cite{touvron2021going}, allowing to remove the need for batch-specific or cross-channel normalizations such as BatchNorm, GroupNorm or LayerNorm.
We speculate that this stability comes from replacing self-attention with linear layers.
Finally, another advantage of using a linear layer is that we can still visualize the interactions between patch embeddings, revealing filters that are similar to convolutions on the lower layers, and longer range in the last layers.

We further investigate if our purely MLP based architecture could benefit to other domains beyond images, and particularly, with more complex output spaces.
In particular, we adapt our MLP based architecture to take inputs with variable length, and show its potential on the problem of Machine Translation.  
To do so, we develop a sequence-to-sequence (seq2seq) version of \OURS, where both encoder and decoders are based on \OURS with across-attention between the encoder and decoder~\cite{bahdanau2014neural}. 
This model is similar to the original seq2seq Transformer with \OURS layers instead of Transformer layers~\cite{vaswani2017attention}.
Despite not being originally designed for this task, we observe that \OURS is competitive with Transformers on the challenging WMT benchmarks.

In summary, in this paper,  we make the following observations:
\begin{itemize}[leftmargin=*]
    \item despite its simplicity, \OURS reaches surprisingly good  accuracy/complexity trade-offs  with \ImNet-1k training only\footnote{Concurrent work by Tolstikhin \etal~\cite{tolstikhin2021MLPMixer} brings complementary insights to ours: they achieve interesting performance with larger MLP models pre-trained on the larger public \ImNet-22k and even more data with the proprietary JFT-300M. In contrast, we focus on faster models trained on \ImNet-1k. 
    Other concurrent related work includes that of Melas-Kyriazi~\cite{melaskyriazi2021doyoueven} and the  RepMLP~\cite{ding2021repmlp} and gMLP~\cite{liu2021mlp} models.}, without 
    requiring normalization  based on batch or channel statistics; 
    \item these models benefit significantly from distillation methods~\cite{Touvron2020TrainingDI}; they are also compatible with modern self-supervised learning methods based on data augmentation, such as  DINO~\cite{caron2021emerging};
    \item A seq2seq \OURS achieves competitive performances compared to a seq2seq Transformers on the WMT benchmark for Machine Translation. 
\end{itemize}

\section{Method}
\label{sec:method}

\begin{figure}
    \centering
     \includegraphics[width=\linewidth]{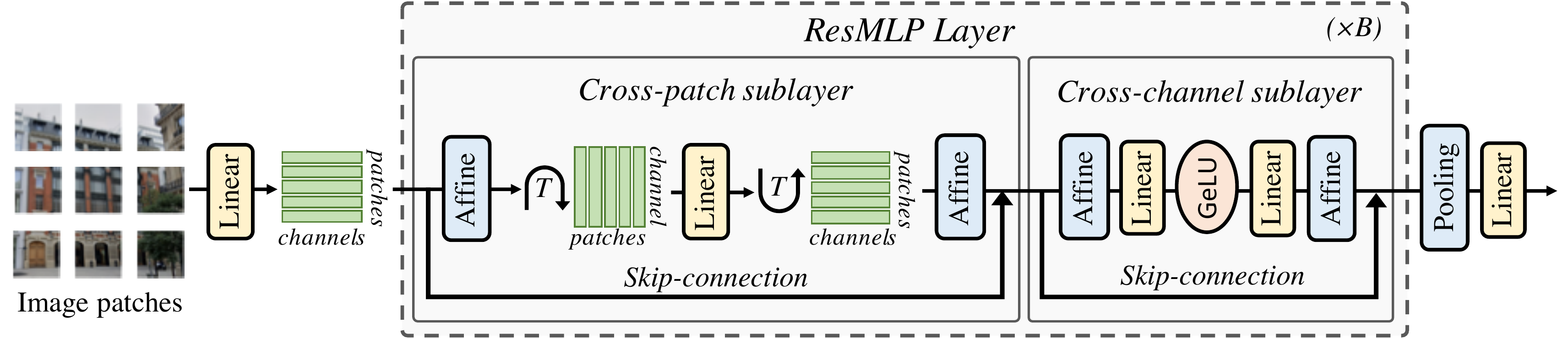} 
    \caption{
    {\bf The \OURS architecture.} 
    After linearly projecting the image patches into high dimensional embeddings, 
    \OURS sequentially processes them with (1) a cross-patch linear sublayer; (2) a cross-channel two-layer MLP. 
    The MLP is the same as the FCN sublayer of a Transformer.
    Each sublayer has a residual connection and two Affine element-wise transformations.
    \label{fig:architecture}}
\end{figure}

In this section, we describe our architecture, \OURS, as depicted in Figure~\ref{fig:architecture}.
\OURS is inspired by ViT and this section focuses on the changes made to ViT that lead to a purely MLP based model. We refer the reader to Dosovitskiy~\etal~\cite{dosovitskiy2020image} for more details about ViT.

\textbf{The overall ResMLP architecture.}
Our model, denoted by ResMLP, takes a grid of $N\!\times\!N$ non-overlapping patches as input, where the patch size is typically equal to $16\!\times\!16$.
The patches are then independently passed through a linear layer to form a set of $N^2$ $d$-dimensional embeddings.

The resulting set of $N^2$ embeddings are fed to a sequence of \emph{Residual Multi-Layer Perceptron} layers to produce a set of $N^2$ $d$-dimensional output embeddings.
These output embeddings are then averaged (``average-pooling'') as a $d$-dimension vector to represent the image, which is fed to a linear classifier to predict the label associated with the image. Training uses the cross-entropy loss.

\textbf{The Residual Multi-Perceptron Layer.}
Our network is a sequence of layers that all have the same structure: a linear sublayer applied across patches followed by a feedforward sublayer applied across channels.
Similar to the Transformer layer, each sublayer is paralleled with a skip-connection~\cite{He2016ResNet}.
The absence of self-attention layers makes the training more stable, allowing us to replace the Layer Normalization~\cite{ba2016layer} by a simpler Affine transformation:
\begin{align}
\texttt{Aff}_{\bm\alpha,\bm\beta}(\mathbf{x}) & = \texttt{Diag} (\bm\alpha) \mathbf{x} + \bm\beta, 
\end{align}
where $\bm \alpha$ and $\bm\beta$ are learnable weight vectors. This operation only rescales and shifts the input element-wise.
This operation has several advantages over other normalization operations:
first, as opposed to Layer Normalization, it has no cost at inference time, since it can absorbed in the adjacent linear layer. %
Second, as opposed to BatchNorm~\cite{Ioffe2015BatchNA} and Layer Normalization, the \texttt{Aff} operator does not depend on batch statistics. 
The closer operator to \texttt{Aff} is the LayerScale introduced by Touvron~\etal~\cite{touvron2021going}, with an additional bias term.
For convenience, we denote by $\texttt{Aff}(\mathbf{X})$ the Affine operation applied independently to each column of the matrix $\mathbf{X}$. 

We apply the \texttt{Aff} operator at the beginning (``pre-normalization'') and end (``post-normalization'') of each residual block. As a pre-normalization,  $\texttt{Aff}$ replaces LayerNorm without using channel-wise statistics. Here, we initialize $\bm \alpha=\bm 1$, and $\bm \beta=\bm 0$.  
As a post-normalization, $\texttt{Aff}$ is similar to LayerScale and we initialize $\bm \alpha$ with the same small value as in~\cite{touvron2021going}. 

Overall, our Multi-layer perceptron takes a set of $N^2$ $d$-dimensional input features stacked in a $d\times N^2$ matrix $\mathbf{X}$, and outputs a set of $N^2$ $d$-dimension output features, stacked in a matrix  $\mathbf{Y}$ with the following set of transformations:
\begin{eqnarray}
\mathbf{Z} &=& \mathbf{X} + \texttt{Aff}\left((\mathbf{A}~\texttt{Aff}\left(\mathbf{X})^\top\right)^\top\right),\label{eq:linear}\\
\mathbf{Y} &=& \mathbf{Z} + \texttt{Aff}\left( \mathbf{C}~\texttt{GELU} (\mathbf{B}~\texttt{Aff}(\mathbf{Z}))\right),\label{eq:linear3}
\end{eqnarray}
where $\mathbf{A}$, $\mathbf{B}$ and $\mathbf{C}$ are the main learnable weight matrices of the layer.
Note that Eq~(\ref{eq:linear3}) is the same as the feedforward sublayer of a Transformer with the \texttt{ReLU} non-linearity replaced by a \texttt{GELU} function~\cite{Hendrycks2016GaussianEL}.
The dimensions of the parameter matrix $\mathbf{A}$ are $N^2\!\times\!N^2$, \ie, this ``cross-patch'' sublayer 
exchanges information between patches, while the ``cross-channel'' feedforward sublayer works per location.
Similar to a Transformer, the intermediate activation matrix $\mathbf{Z}$ has the same dimensions as the input and output matrices, $\mathbf{X}$ and $\mathbf{Y}$.
Finally, the weight matrices $\mathbf{B}$ and $\mathbf{C}$ have the same dimensions as in a Transformer layer, which are $4d\!\times\!d$ and $d\!\times\!4d$, respectively. 

\textbf{Differences with the Vision Transformer architecture.} Our architecture is closely related to the ViT model~\cite{dosovitskiy2020image}. However, \OURS departs from ViT with several simplifications: 
\begin{itemize}[leftmargin=*]
    \item \emph{no self-attention blocks}: it is replaced by a linear layer with no  non-linearity,
    \item \emph{no positional embedding}: the linear layer implicitly encodes information about patch positions,
    \item \emph{no extra ``class'' token}: we simply use average pooling on the patch embeddings,
    \item \emph{no normalization based on batch statistics}: we use a learnable affine operator. 
\end{itemize}

\textbf{Class-MLP as an alternative to average-pooling.} 
We propose an adaptation of the class-attention token introduced in  CaiT~\cite{touvron2021going}. In CaiT, this consists of two layers that have the same structure as the transformer, but in which only the class token is updated based on the frozen  patch embeddings.  
We translate this method to our architecture, except that, after aggregating the patches with a linear layer, we replace the attention-based interaction between the class and patch embeddings by simple linear layers, still keeping the patch embeddings frozen.
This increases the performance, at the expense of adding some parameters and computational cost. 
We refer to this pooling variant as ``class-MLP'', since the purpose of these few layers is to replace average pooling.

\textbf{Sequence-to-sequence ResMLP.}
Similar to Transformer, the ResMLP architecture can be applied to sequence-to-sequence tasks. First, we follow the general encoder-decoder architecture from Vaswani \etal~\cite{vaswani2017attention}, where we replace the self-attention sublayers by the residual multi-perceptron layer. In the decoder, we keep the cross-attention sublayers, which attend to the output of the encoder. 
In the decoder, we adapt the linear sublayers to the task of language modeling by constraining the matrix $\mathbf{A}$ to be triangular, in order to prevent a given token representation to access tokens from the future.
Finally, the main technical difficulty from using linear sublayers in a sequence-to-sequence model is to deal with variable sequence lengths.
However, we observe that simply padding with zeros and extracting the submatrix $\mathbf{A}$ corresponding to the longest sequence in a batch, works well in practice. 

\section{Experiments}

In this section, we present experimental results for the \OURS architecture on image  classification and machine translation. 
We also study the impact of the different components of \OURS in ablation studies.
We consider three training paradigms for images:
\begin{itemize}[leftmargin=*]
\item
\emph{Supervised learning:} We train \OURS  from labeled images with a softmax classifier  and cross-entropy loss. This paradigm is the main focus of our work. 
\item \emph{Self-supervised learning:} We train the \OURS with the DINO method of Caron~\etal~\cite{caron2021emerging} that trains a network without labels by distilling knowledge from previous instances of the same network.
\item \emph{Knowledge distillation:} We employ the knowledge distillation procedure proposed by Touvron \etal~\cite{Touvron2020TrainingDI} to guide the supervised training of \OURS with a convnet. 
\end{itemize}

\subsection{Experimental setting}
\label{sec:experimental_setting}

\textbf{Datasets.} 
We train our models on the \ImNet-1k dataset~\cite{Russakovsky2015ImageNet12}, that contains 1.2M images evenly spread over 1,000 object categories. 
In the absence of an available test set for this benchmark, we follow the standard practice in the community by reporting performance on the validation set.
This is not ideal since the validation set was originally designed to select hyper-parameters.
Comparing methods on this set may not be conclusive enough because an improvement in performance may not be caused by better modeling, but by a better selection of hyper-parameters.
To mitigate this risk, we report additional results in transfer learning and on two alternative versions of \ImNet that have been built to have distinct validation and test sets, namely the \ImNet-real~\cite{Beyer2020ImageNetReal} and \ImNet-v2~\cite{Recht2019ImageNetv2} datasets. %
We also report a few data-points when training on \ImNet-21k. 
Our hyper-parameters are mostly adopted from Touvron et al.~\cite{Touvron2020TrainingDI,touvron2021going}.

\textbf{Hyper-parameter settings.}
In the case of supervised learning, we train our network with the Lamb optimizer~\cite{you20lamb} with a learning rate of $5 \times 10^{-3}$ and weight decay $0.2$.
We initialize the LayerScale parameters as a function of the depth by following %
CaiT~\cite{touvron2021going}.
The rest of the hyper-parameters follow the default setting used in DeiT~\cite{Touvron2020TrainingDI}.
For the knowledge distillation paradigm, we use the same RegNety-16GF~\cite{Radosavovic2020RegNet} as in DeiT with the same training schedule.
The majority of our models take two days to train on eight V100-32GB GPUs.%

\begin{table}[t]

    \caption{
\textbf{Comparison between architectures on \ImNet  classification.}
We compare different architectures  based on convolutional networks, Transformers and feedforward networks with comparable FLOPs and number of parameters. 
We report Top-1 accuracy on the validation set of \ImNet-1k with different measure of complexity: throughput, FLOPs, number of parameters and peak memory usage.
All the models use $224\!\times\!224$ images as input.
By default the Transformers and feedforward networks uses $14\!\times\!14$ patches of size $16\!\times\!16$, 
see Table~\ref{tab:ablation} for the detailed specification of our main models. 
The throughput is measured on a single V100-32GB GPU with batch size fixed to 32. 
For reference, we include the state of the art with \ImNet training only. 
    \label{tab:mainres}}
    \centering
    \smallskip
    \scalebox{0.85}{
    \begin{tabular}{l|lcccccc}
        \toprule
        &  Arch.        & \#params & throughput & FLOPS & Peak Mem & Top-1\\
         &             & ($\times 10^6$) & (im/s) & ($\times 10^9$) & (MB) & Acc.\\
        \midrule
        \multirow{2}{*}{{\emph{State of the art}}} &
        CaiT-M48$\uparrow$448$\Upsilon$~\cite{touvron2021going} & 356\pzo  & \dzo5.4 & \pzo329.6 & 5477.8 & 86.5  \\
        & NfNet-F6 SAM~\cite{Brock2021HighPerformanceLI}   & 438\pzo & \pzo16.0 & \pzo377.3& 5519.3& 86.5  \\
        \midrule
	\multirow{7}{*}{\emph{Convolutional networks}} 
	& EfficientNet-B3~\cite{tan2019efficientnet} & 12  & 661.8 & \tzo1.8  & 1174.0    & 81.1\\
	& EfficientNet-B4~\cite{tan2019efficientnet} & 19  & 349.4 & \tzo4.2 & 1898.9    & 82.6\\
	& EfficientNet-B5~\cite{tan2019efficientnet} & 30  & 169.1 & \tzo9.9 & 2734.9    & 83.3\\
	& RegNetY-4GF~\cite{radenovic2018fine}       & 21  & 861.0 & \tzo4.0 & \pzo568.4 & 80.0\\
	& RegNetY-8GF~\cite{radenovic2018fine}       & 39  & 534.4 & \tzo8.0 & \pzo841.6 & 81.7\\
	& RegNetY-16GF~\cite{radenovic2018fine}      & 84  & 334.7 & \dzo16.0 & 1329.6    & 82.9\\
        \midrule
       \multirow{3}{*}{\emph{Transformer networks}}   
         & DeiT-S~\cite{Touvron2020TrainingDI}   & 22  & 940.4 & \tzo4.6 & \pzo217.2 & 79.8\\
	 & DeiT-B~\cite{Touvron2020TrainingDI}       & 86  & 292.3 & \dzo17.5 & \pzo573.7 & 81.8\\
	 & CaiT-XS24~\cite{touvron2021going}         & 27  & 447.6 & \tzo5.4 & \pzo245.5 & 81.8\\
        \midrule
\multirow{3}{*}{\emph{Feedforward networks}}   
        & \OURS-S12 &  15  & 1415.1\pzo & \tzo3.0  & \pzo179.5 &  76.6 \\
        & \OURS-S24 &  30  & 715.4      & \tzo6.0  & \pzo235.3 &  79.4 \\
        & \OURS-B24 &  116\pzo &   231.3         & \dzo23.0 &  \pzo663.0 &   81.0 \\
        \bottomrule
    \end{tabular}}
\end{table}

\subsection{Main Results}

In this section, we compare \OURS with architectures based on convolutions or self-attentions with comparable size and throughput on \ImNet.

\textbf{Supervised setting.}
In Table~\ref{tab:mainres}, we compare \OURS with different convolutional and Transformer architectures. 
For completeness, we also report the best-published numbers obtained with a model trained on \ImNet alone. 
While the trade-off between accuracy, FLOPs, and throughput for \OURS is not as good as convolutional networks or Transformers, their strong accuracy still suggests that the structural constraints imposed by the layer design do not have a drastic influence on performance, especially when training with enough data and recent training schemes.

\begin{table}[t]
    \captionof{table}{\textbf{Self-supervised learning} with DINO~\cite{caron2021emerging}. 
    Classification accuracy on \ImNet-1k val.
    ResMLPs evaluated with linear and $k$-NN evaluation on ImageNet are comparable to convnets but inferior to ViT.  \smallskip } 
    
    \def \mysp {\hspace{6pt}}
     \centering \scalebox{0.85}
    {
    \begin{tabular}{lcccccc}
    \toprule
    Models                   &  ResNet-50 &  ViT-S/16 & ViT-S/8 &  ViT-B/16 & \OURS-S12 &   \OURS-S24 \\
    \midrule
    Params. ($\times 10^6$) &   25        & 22       &  22     &  87      & 15       & 30\\
    FLOPS ($\times 10^9$)   &  4.1        &  4.6     & 22.4    & 17.5     & 3.0      & 6.0\\
    \midrule
    Linear                  &   75.3      & 77.0     &  79.7   & 78.2     &    67.5      & 72.8\\
    $k$-NN                  & 67.5         & 74.5    & 78.3   & 76.1      &   62.6       &  69.4\\
    \bottomrule
    \end{tabular}}  
    \label{tab:ssl}
\end{table}

\textbf{Self-supervised setting. } 
We pre-train \OURS-S12 using the self-supervised method called DINO~\cite{caron2021emerging} during 300 epochs. 
We report our results in Table~\ref{tab:ssl}.
The trend is similar to the supervised setting: the accuracy obtained with \OURS is lower than ViT. Nevertheless, the performance is surprisingly high for a pure MLP architecture and competitive with Convnet in $k$-NN evaluation. 
Additionally, we also fine-tune network pre-trained with self-supervision on \ImNet using the ground-truth labels.
Pre-training substantially improves performance compared to a \OURS-S24 solely trained with labels, achieving 79.9\% top-1 accuracy on \ImNet-val (+0.5\%). 

\textbf{Knowledge distillation setting.}
We study our model when training with the knowledge distillation approach of Touvron~\etal~\cite{Touvron2020TrainingDI}.
In their work, the authors show the impact of training a ViT model by distilling it from a RegNet. 
In this experiment, we explore if \OURS also benefits from this procedure and summarize our results in Table~\ref{tab:ablation} (Blocks ``Baseline models'' and  ``Training'').
We observe that similar to DeiT models, \OURS greatly benefits from distilling from a convnet. 
This result concurs with the observations made by d'Ascoli~\etal~\cite{d2019finding}, who used convnets to initialize feedforward networks.
Even though our setting differs from theirs in scale, the problem of overfitting for feedforward networks is still present on \ImNet. 
The additional regularization obtained from the distillation is a possible explanation for this improvement.

\subsection{Visualization \& analysis of the linear interaction between patches}

\begin{figure}
\newcommand{\myvisu}[1]{{ \includegraphics[width=.237\linewidth]{figs/FilterVisu/layer_#1_w_fc1.pdf}}}
\small
\hspace{-1pt}
\begin{tabular}{@{}c@{\ \ }c@{\ \ }c@{\ \ }c@{\ \ }@{}}
       \myvisu{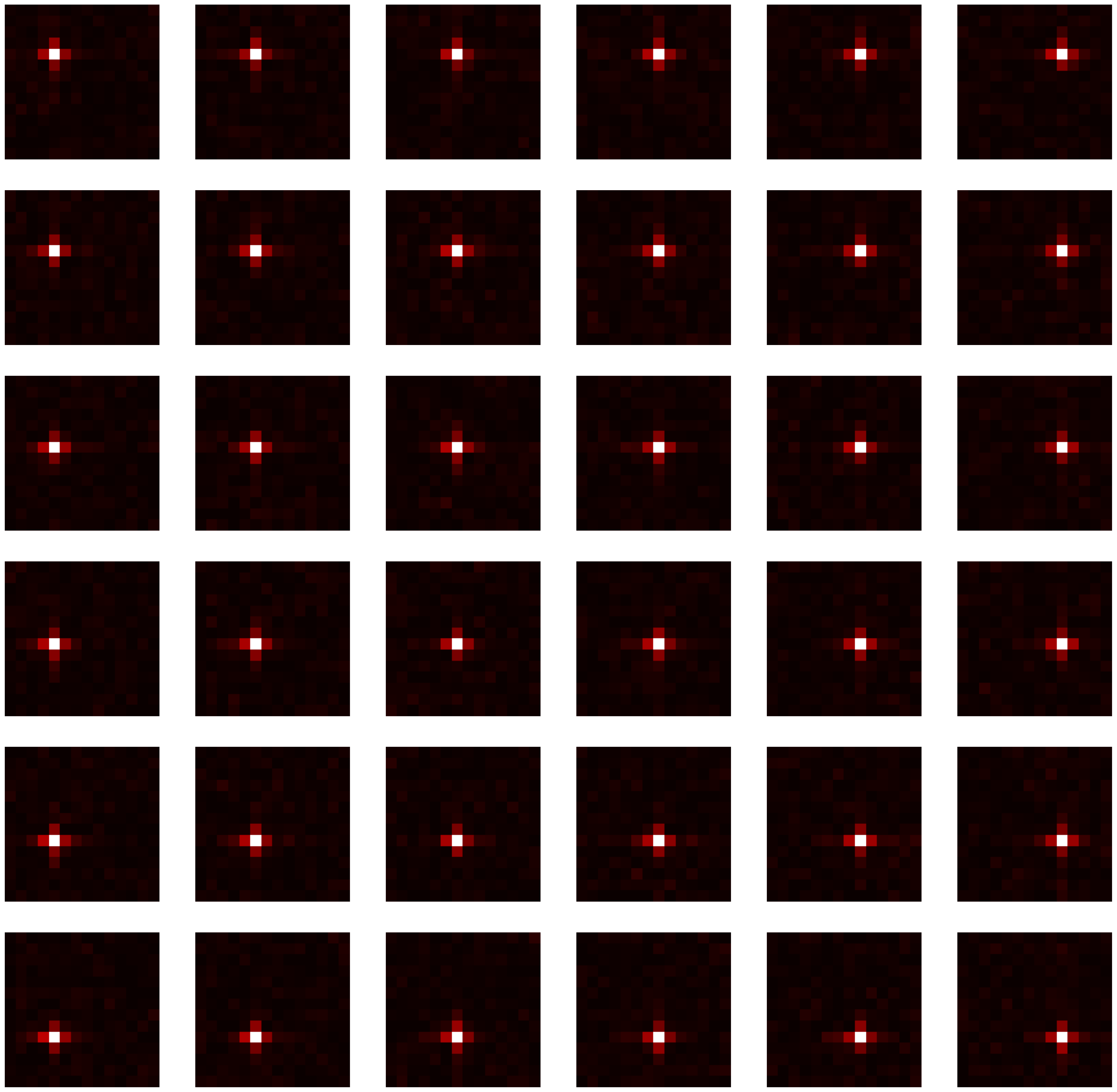} &  \myvisu{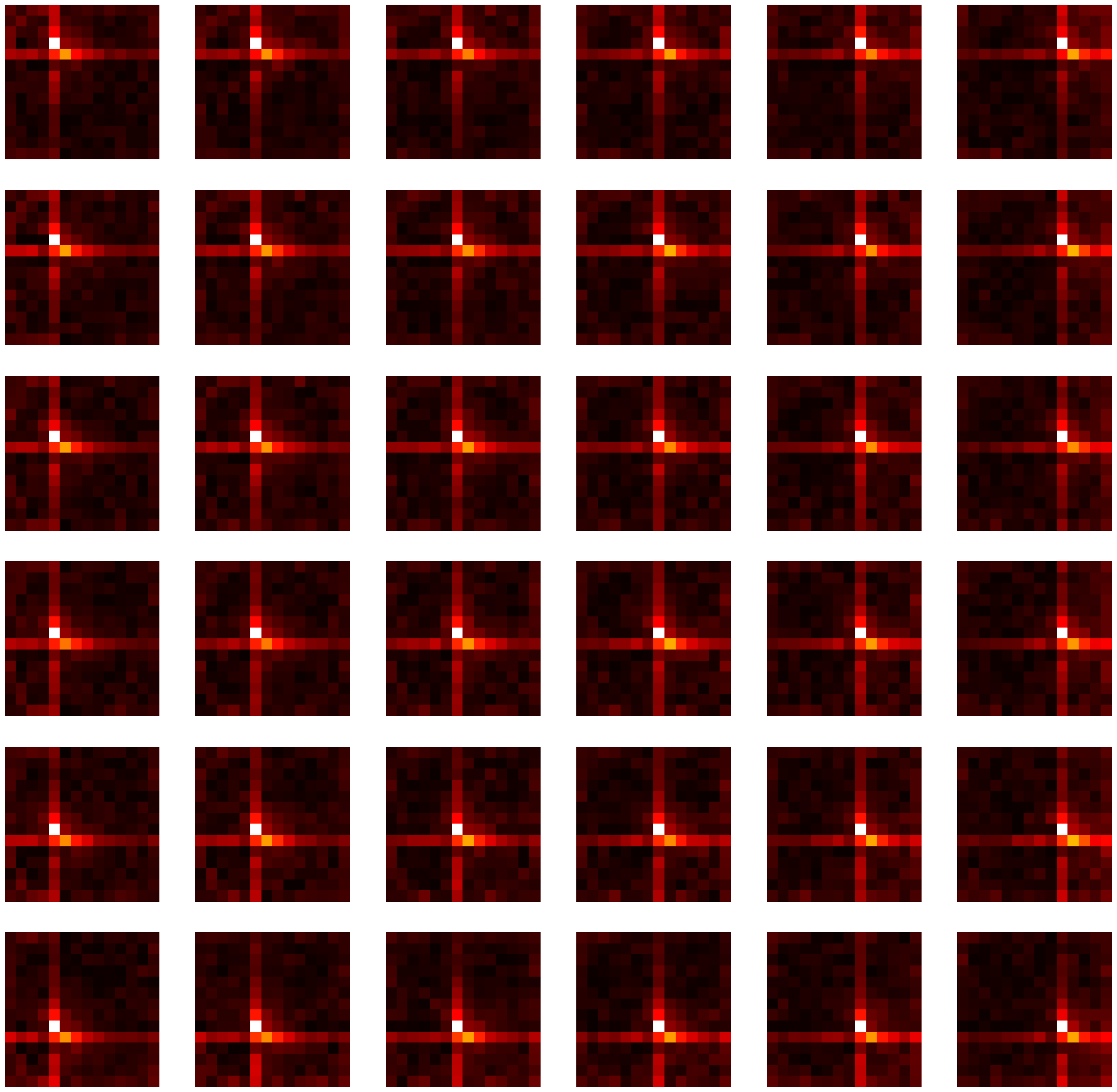}  &  \myvisu{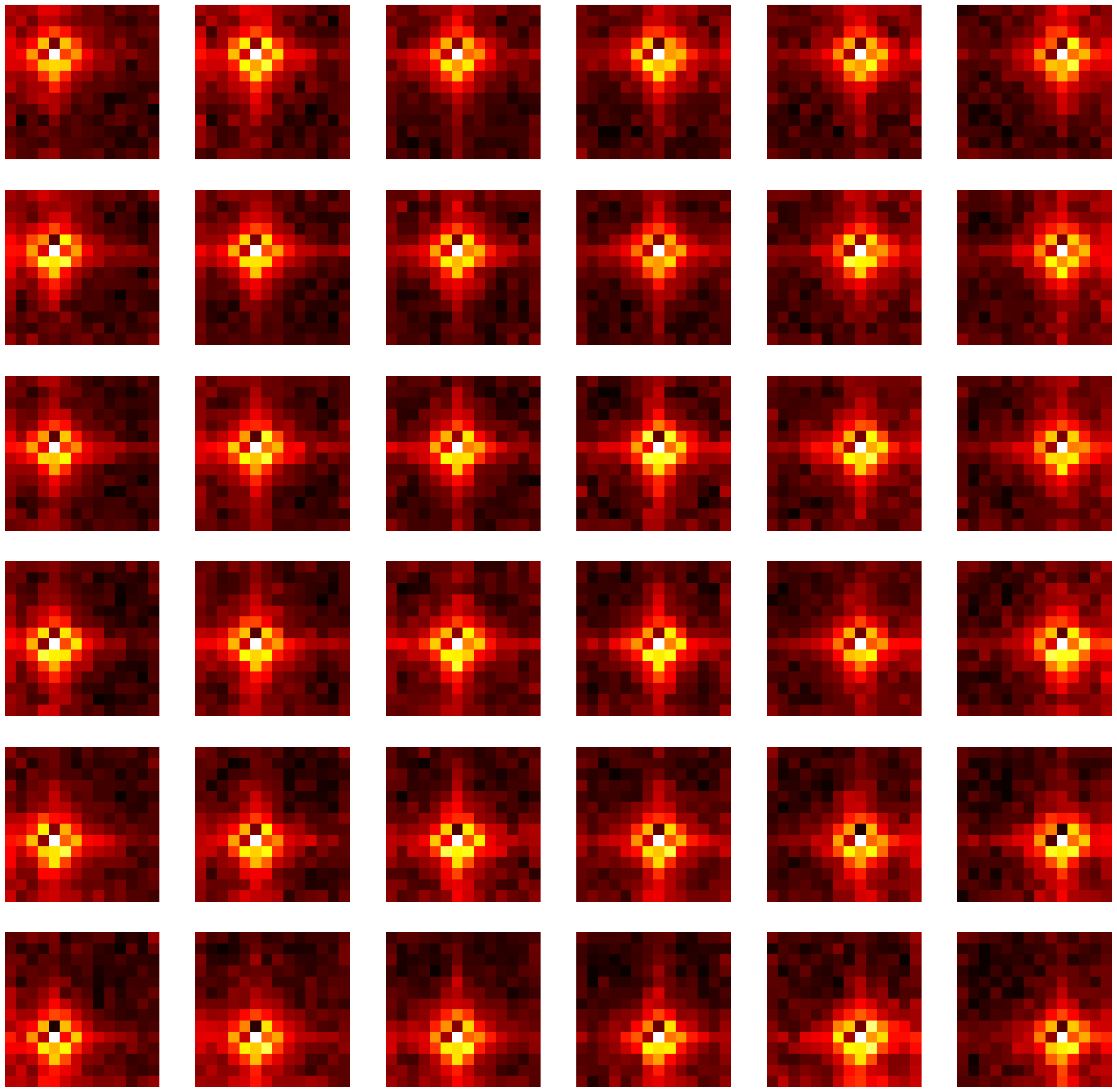} & \myvisu{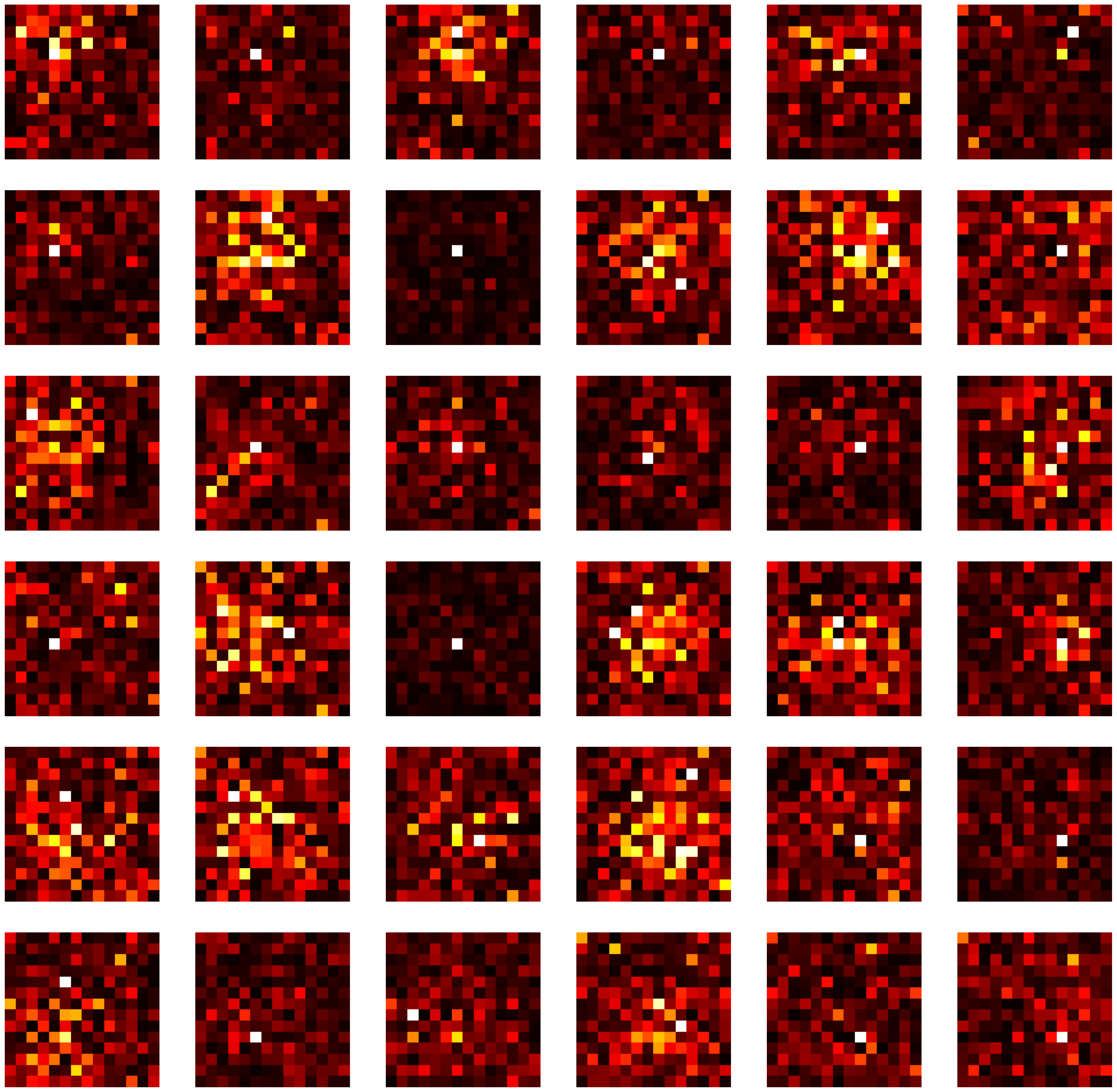} \\
       Layer 1 & Layer 7  &  Layer 10 & Layer 20 
\end{tabular}
 \caption{
{\bf Visualisation of the linear layers in \OURS-S24.} 
For each layer we visualise the rows of the matrix $\mathbf{A}$ as a set of $14\times 14$ pixel images, for sake of space we only show the rows corresponding to the 6$\times$6 central patches. 
We observe patterns in the linear layers that share similarities with convolutions. 
In appendix~\ref{sec:vizu_unsup} we provide comparable visualizations for all layers of a ResMLP-S12 model. 
\label{fig:vizu}}
\end{figure}

\textbf{Visualisations of the cross-patch sublayers.}
In Figure~\ref{fig:vizu}, we show in the form of squared images, the rows of the weight matrix from cross-patch sublayers at different depths of a \OURS-S24 model. 
The early layers show convolution-like patterns: the weights resemble shifted versions of each other and have local support. 
Interestingly, in many layers, the support also extends along both axes; see layer 7.
The last 7 layers of the network are different: they consist of a spike for the patch itself and a diffuse response across other patches with different magnitude; see layer 20.

\begin{figure}
\begin{minipage}{0.48\linewidth}
    \centering \includegraphics[width=0.99\linewidth]{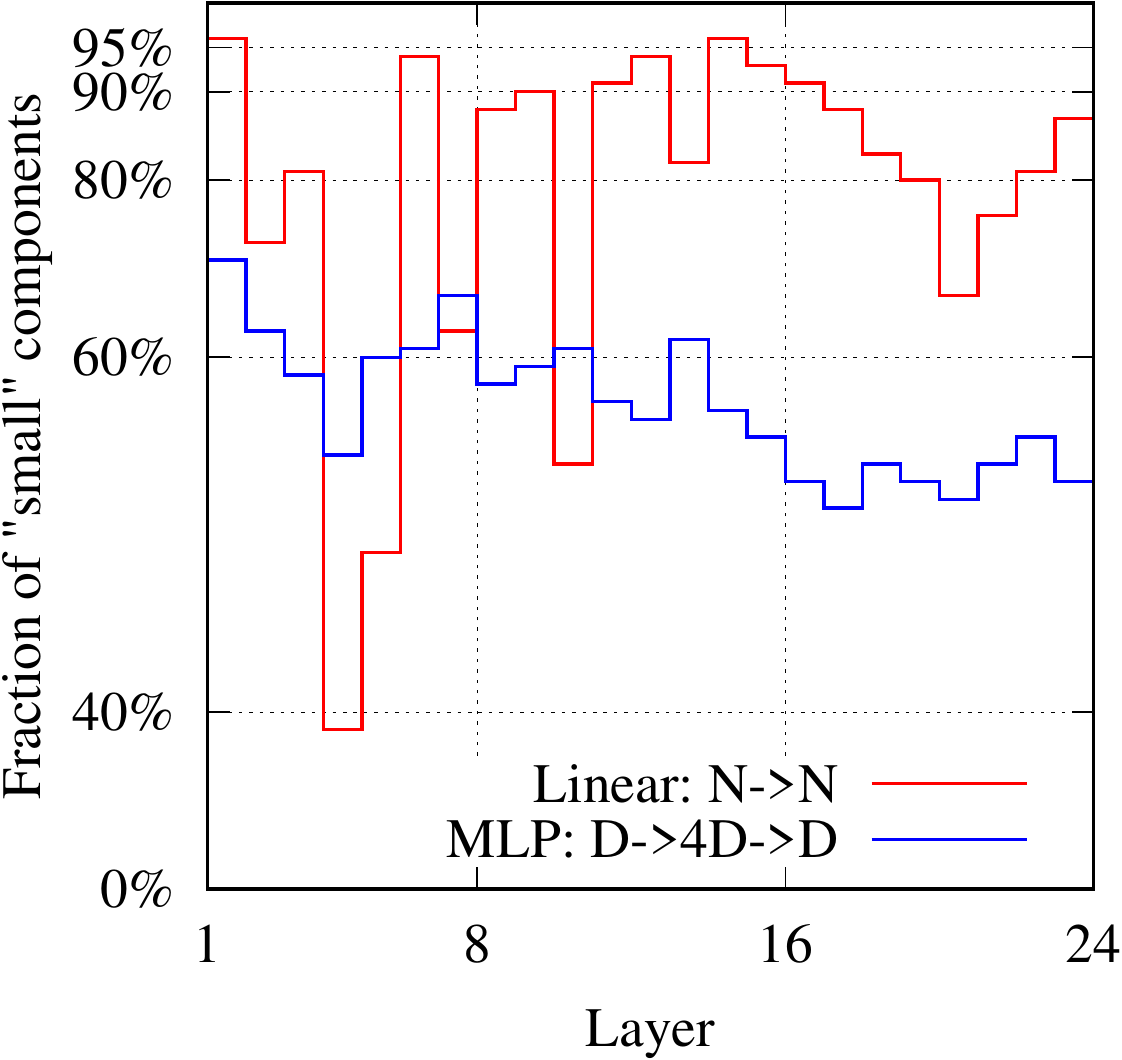}
    \caption{%
    \textbf{Sparsity of linear interaction layers.}
    For each layer (\textcolor{red!80}{linear} and \textcolor{blue!80}{MLP}), we show the rate of components whose absolute value is lower than 5\% of the maximum. 
    Linear interaction layers  are   sparser than the matrices involved in the per-patch  MLP. 
    \label{fig:sparsity}}
\end{minipage}
\hfill
\begin{minipage}{0.48\linewidth}
    \centering \includegraphics[width=0.99\linewidth]{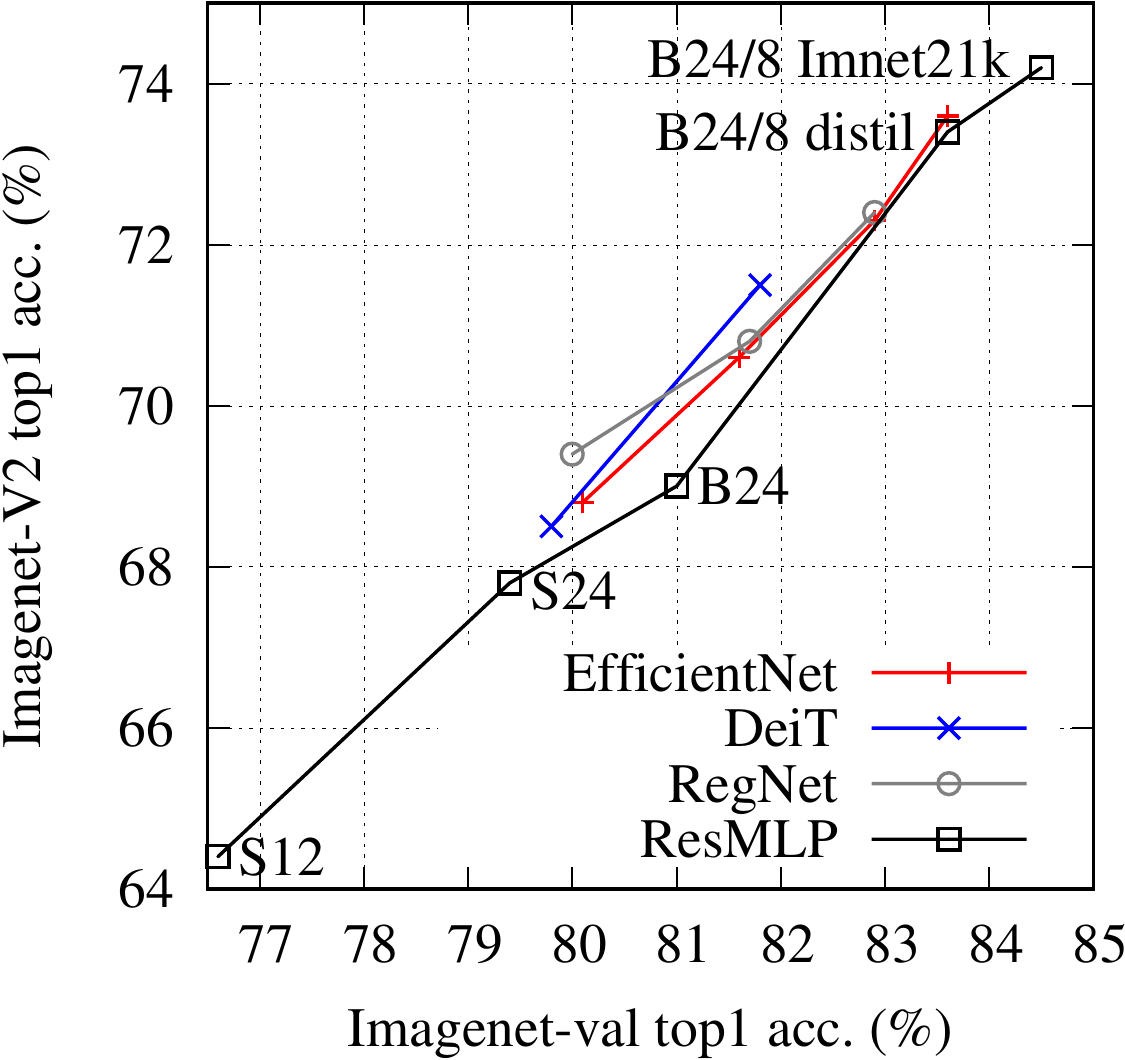}
    \caption{
    \textbf{Top-1 accuracy on \ImNet-V2 \vs  \ImNet-val.} 
    ResMLPs tend to overfit slightly more under identical training method. This is partially alleviated with by introducing more regularization (more data or distillation, see e.g., \OURS-B24/8-distil). %
    \label{fig:overfitanalysis}}    
\end{minipage}
\end{figure}

\textbf{Measuring sparsity of the weights.} 
The visualizations described above suggest that the linear communication layers are sparse. 
We analyze this quantitatively in more detail in Figure~\ref{fig:sparsity}. 
We measure the sparsity of the matrix $\mathbf{ A}$, and compare it to the sparsity of $\mathbf{ B}$ and $\mathbf{ C}$ from the per-patch MLP. 
Since there are no exact zeros, we measure the rate of components whose absolute value is lower than 5\% of the maximum value. 
Note, discarding the small values is analogous to the case where we normalize the matrix by its maximum and use a finite-precision representation of weights. 
For instance, with a 4-bits representation of weight, one would typically round to zero all weights whose absolute value is below 6.25\% of the maximum value. 

The measurements in Figure~\ref{fig:sparsity} show that all three matrices are sparse, with the layers implementing the patch communication being significantly more so. 
This suggests that they may be compatible with parameter pruning, or better, with modern quantization techniques that induce sparsity at training time, such as Quant-Noise~\cite{fan2020training} and DiffQ~\cite{defossez2021differentiable}. 
The sparsity structure, in particular in earlier layers, see Figure.~\ref{fig:vizu}, hints that we could implement the patch interaction linear layer with a convolution. 
We provide some results for convolutional variants in our ablation study. Further research on network compression is beyond the scope of this paper, yet we believe it worth investigating in the future. 

\textbf{Communication across patches } 
if we remove the linear interaction layer (linear $\rightarrow$ none), we obtain substantially lower accuracy (-20\% top-1 acc.) for a ``bag-of-patches'' approach. 
We have tried several alternatives for the cross-patch sublayer, which are presented in Table~\ref{tab:ablation} (block ``patch communication''). 
Amongst them, using the same MLP structure as for patch processing (linear $\rightarrow$ MLP), which we analyze in more details in the supplementary material. 
The simpler choice of a single linear square layer led to a better accuracy/performance trade-off -- considering that the MLP variant requires compute halfway between \OURS-S12 and \OURS-S24 -- and requires fewer parameters than a residual MLP block. 

The  visualization in Figure \ref{fig:vizu} indicates that many linear interaction layers %
look like convolutions. 
In our ablation, we  replaced the linear layer with different types of $3\!\times\!3$ convolutions. %
The depth-wise convolution does not implement interaction across channels  -- as  our linear patch communication layer --  and yields similar performance at a comparable number of parameters and FLOPs.
While  full $3\!\times\!3$ convolutions yield best results, they come with roughly double the number of parameters and FLOPs. 
Interestingly, the depth-separable convolutions combine accuracy close to that of full $3\!\times\!3$ convolutions with  a  number of parameters and FLOPs comparable to our linear layer. 
This suggests that  convolutions on low-resolution feature maps at all layers is an interesting alternative to the  common pyramidal design of convnets, where early layers operate at higher resolution and smaller feature dimension.

\begin{table}[]
    \small \centering
    \scalebox{0.79}{
    \begin{tabular}{l|l@{\ \ }cc|c|l|cccc}
    \toprule
\multirow{2}{*}{Ablation}  &  \multirow{2}{*}{Model} & Patch & Params & FLOPs & \multirow{2}{*}{Variant} &  \multicolumn{3}{c}{top-1 acc. on \ImNet} \\
  &  & size & $\times 10^6$ & $\times 10^9$ & & val & real~\cite{Beyer2020ImageNetReal} & v2~\cite{Recht2019ImageNetv2} \\
  \midrule
  & \cellcolor{yellow!5} \OURS-S12 &  \cellcolor{yellow!5} 16 & \cellcolor{yellow!5} 15.4  & \cellcolor{yellow!5} 3.0  &  \cellcolor{yellow!5} 12 layers, working dimension 384  & \cellcolor{yellow!5} 76.6 &  \cellcolor{yellow!5} 83.3 & \cellcolor{yellow!5} 64.4 \\ 
Baseline models & \cellcolor{red!5}  \OURS-S24 & \cellcolor{red!5} 16 & \cellcolor{red!5} 30.0  & \cellcolor{red!5} 6.0  &  \cellcolor{red!5} 24 layers, working dimension 384    & \cellcolor{red!5} 79.4 & \cellcolor{red!5} 85.3 & \cellcolor{red!5} 67.9  \\ 
  &  \cellcolor{blue!3}\OURS-B24 & \cellcolor{blue!3} 16 & \cellcolor{blue!3} 115.7\pzo &\cellcolor{blue!3}  23.0\pzo &  \cellcolor{blue!3} 24 layers, working dimension 768    & \cellcolor{blue!3} 81.0 &  \cellcolor{blue!3} 86.1 & \cellcolor{blue!3} 69.0 \\
  \midrule
\multirow{1}{*}{Normalization}
  &  \cellcolor{yellow!5}\OURS-S12 & \cellcolor{yellow!5} 16 & \cellcolor{yellow!5} 15.4 & \cellcolor{yellow!5} 3.0 &  \cellcolor{yellow!5} \texttt{Aff} $\rightarrow$ Layernorm & \cellcolor{yellow!5} 77.7 & \cellcolor{yellow!5} 84.1 & \cellcolor{yellow!5} 65.7  \\ 
  \midrule
\multirow{1}{*}{Pooling}
  &  \cellcolor{yellow!5}\OURS-S12 & \cellcolor{yellow!5} 16 & \cellcolor{yellow!5} 17.7 & \cellcolor{yellow!5} 3.0 & \cellcolor{yellow!5} average pooling $\rightarrow$ Class-MLP & \cellcolor{yellow!5} 77.5 & \cellcolor{yellow!5} 84.0 & \cellcolor{yellow!5} 66.1 \\
  \midrule
\multirow{4}{1.9cm}{\begin{minipage}{1.4cm}Patch\\communication\ \end{minipage}}
  &  \cellcolor{yellow!5}\OURS-S12   & \cellcolor{yellow!5}16 & \cellcolor{yellow!5}14.9 & \cellcolor{yellow!5}2.8  & \cellcolor{yellow!5}linear $\rightarrow$ none &\cellcolor{yellow!5} 56.5  &  \cellcolor{yellow!5} 63.4 & \cellcolor{yellow!5} 43.1 \\ 
  &  \cellcolor{yellow!5}\OURS-S12   & \cellcolor{yellow!5} 16 & \cellcolor{yellow!5} 18.6  & \cellcolor{yellow!5} 4.3  & \cellcolor{yellow!5} linear $\rightarrow$ MLP  & \cellcolor{yellow!5} 77.3  & \cellcolor{yellow!5}  84.0 & \cellcolor{yellow!5} 65.7 \\ 
  &  \cellcolor{yellow!5}\OURS-S12   & \cellcolor{yellow!5} 16 & \cellcolor{yellow!5} 30.8 & \cellcolor{yellow!5} 6.0  & \cellcolor{yellow!5} linear $\rightarrow$  conv 3x3 & \cellcolor{yellow!5} 77.3 & \cellcolor{yellow!5} 84.4 & \cellcolor{yellow!5} 65.7\\ 
  &  \cellcolor{yellow!5}\OURS-S12   &\cellcolor{yellow!5}  16 & \cellcolor{yellow!5} 14.9 & \cellcolor{yellow!5} 2.8  & \cellcolor{yellow!5} linear $\rightarrow$ conv 3x3 depth-wise & \cellcolor{yellow!5} 76.3 & \cellcolor{yellow!5} 83.4 & \cellcolor{yellow!5} 64.6 \\ 
  &  \cellcolor{yellow!5}\OURS-S12   &  \cellcolor{yellow!5} 16 &  \cellcolor{yellow!5} 16.7 &  \cellcolor{yellow!5} 3.2  &  \cellcolor{yellow!5} linear $\rightarrow$ conv 3x3 depth-separable &  \cellcolor{yellow!5} 77.0 &  \cellcolor{yellow!5} 84.0 & \cellcolor{yellow!5}  65.5\\ 
  \midrule
\multirow{3}{*}{Patch size}  
  &   \cellcolor{yellow!5}\OURS-S12/14\,  &  \cellcolor{yellow!5} 14 &  \cellcolor{yellow!5} 15.6 &  \cellcolor{yellow!5} 4.0   &  \cellcolor{yellow!5} patch size 16$\times$16$\rightarrow$14$\times$14 &  \cellcolor{yellow!5} 76.9 &  \cellcolor{yellow!5} 83.7 &  \cellcolor{yellow!5} 65.0  \\ 
  &   \cellcolor{yellow!5}\OURS-S12/8  &  \cellcolor{yellow!5} 8  &  \cellcolor{yellow!5} 22.1 &  \cellcolor{yellow!5} 14.0\pzo   & \cellcolor{yellow!5}  patch size 16$\times$16$\rightarrow$8$\times$8 &  \cellcolor{yellow!5} 79.1 &  \cellcolor{yellow!5} 85.2 &  \cellcolor{yellow!5} 67.2  \\ 
  &  \cellcolor{blue!3}\OURS-B24/8    & \cellcolor{blue!3} 8  & \cellcolor{blue!3} 129.1\pzo & \cellcolor{blue!3} 100.2\dzo & \cellcolor{blue!3} patch size 16$\times$16$\rightarrow$8$\times$8 & \cellcolor{blue!3} 81.0 &  \cellcolor{blue!3} 85.7 &\cellcolor{blue!3}  68.6 \\ 
  \midrule
\multirow{7}{*}{Training}  
  &  \cellcolor{yellow!5}\OURS-S12 & \cellcolor{yellow!5} 16 & \cellcolor{yellow!5} 15.4 & \cellcolor{yellow!5} 3.0 & \cellcolor{yellow!5} old-fashioned (90 epochs)   &  \cellcolor{yellow!5} 69.2 & \cellcolor{yellow!5} 76.0 & \cellcolor{yellow!5} 56.1 \\ 
  &  \cellcolor{yellow!5}\OURS-S12 & \cellcolor{yellow!5} 16 & \cellcolor{yellow!5} 15.4 & \cellcolor{yellow!5} 3.0 & \cellcolor{yellow!5} pre-trained SSL (DINO)    &\cellcolor{yellow!5} 76.5  & \cellcolor{yellow!5} 83.6 & \cellcolor{yellow!5} 64.5 \\ 
  &  \cellcolor{yellow!5}\OURS-S12 & \cellcolor{yellow!5} 16 & \cellcolor{yellow!5} 15.4 &\cellcolor{yellow!5} 3.0 & \cellcolor{yellow!5} distillation               &  \cellcolor{yellow!5} 77.8 & \cellcolor{yellow!5} 84.6 & \cellcolor{yellow!5} 66.0 \\ 
  &  \cellcolor{red!5} \OURS-S24 & \cellcolor{red!5} 16 & \cellcolor{red!5} 30.0 & \cellcolor{red!5} 6.0 & \cellcolor{red!5} pre-trained SSL (DINO)       &  \cellcolor{red!5} 79.9 & \cellcolor{red!5} 85.9 & \cellcolor{red!5} 68.6 \\ 
  &  \cellcolor{red!5} \OURS-S24 & \cellcolor{red!5} 16 & \cellcolor{red!5} 30.0 & \cellcolor{red!5} 6.0 & \cellcolor{red!5} distillation               &  \cellcolor{red!5} 80.8 & \cellcolor{red!5} 86.6 & \cellcolor{red!5} 69.8 \\ 
  &  \cellcolor{blue!3}\OURS-B24/8 & \cellcolor{blue!3} 8 & \cellcolor{blue!3} 129.1\pzo & \cellcolor{blue!3} 100.2\dzo & \cellcolor{blue!3} distillation     &  \cellcolor{blue!3} 83.6 & \cellcolor{blue!3} 88.4 & \cellcolor{blue!3} 73.4  \\ 
  &  \cellcolor{blue!3}\OURS-B24/8 & \cellcolor{blue!3} 8  &  \cellcolor{blue!3} 129.1\pzo &  \cellcolor{blue!3} 100.2\dzo &  \cellcolor{blue!3} pre-trained \ImNet-21k (60 epochs)  & \cellcolor{blue!3} {84.4} & \cellcolor{blue!3} {88.9} & \cellcolor{blue!3} {74.2}  \\ 
    \bottomrule
\end{tabular}}
    \smallskip
    \caption{\textbf{Ablation.} Our default configurations are presented in the three first rows. By default we train during 400 epochs. The ``old-fashioned'' is similar to what was employed for ResNet~\cite{He2016ResNet}: SGD, 90-epochs waterfall schedule, same augmentations up to variations due to  library used.\label{tab:ablation}}
\end{table}

\subsection{Ablation studies}
Table~\ref{tab:ablation} reports the ablation study of our base network and a summary of our preliminary exploratory studies. 
We discuss the ablation below and give more detail about early experiments in Appendix~\ref{sec:exploration}. 

\textbf{Control of overfitting.} 
Since MLPs are subject to overfitting, we show in Fig.~\ref{fig:overfitanalysis} a control experiment to probe for problems with generalization.
We explicitly analyze the differential of performance between the \ImNet-val and the distinct \ImNet-V2 test set. 
The relative offsets between curves reflect to which extent models are overfitted to \ImNet-val \wrt hyper-parameter selection. 
The degree of overfitting of our MLP-based model is overall neutral or slightly higher to that of other transformer-based architectures or convnets with same training procedure.

\textbf{Normalization \& activation.}
Our network configuration does not contain any batch normalizations. 
Instead, we use the  affine per-channel transform \texttt{Aff}. 
This is akin to Layer Normalization~\cite{ba2016layer}, typically used in transformers, except that we avoid to collect any sort of statistics, since we do no need it it for convergence. 
In preliminary experiments with pre-norm and post-norm ~\cite{He2016IdentityMappings}, we observed that both choices converged.
Pre-normalization in conjunction with Batch Normalization could provide an accuracy gain in some cases, see Appendix~\ref{sec:exploration}.

We choose to use a GELU~\cite{Hendrycks2016GaussianEL} function. In Appendix~\ref{sec:exploration} we also analyze the activation function: 
ReLU~\cite{glorot11relu} also gives a good performance, but it was a bit more unstable in some settings. 
We did not manage to get good results with SiLU~\cite{Hendrycks2016GaussianEL} and HardSwish~\cite{Howard2019SearchingFM}. 

\textbf{Pooling.}  Replacing average pooling with Class-MLP, see Section~\ref{sec:method}, brings a significant gain for a negligible computational cost. We do not include it by default to keep our models more simple.

\textbf{Patch size.} 
Smaller patches significantly increase the performance, but also increase the number of flops (see Block "Patch size" in Table~\ref{tab:ablation}).
Smaller patches benefit more to larger models, but only with an improved optimization scheme involving more regularization (distillation) or more data. 

\textbf{Training.} Consider the Block ``Training' in Table~\ref{tab:ablation}. %
ResMLP significantly benefits  from modern training procedures such as those used in DeiT. 
For instance, the DeiT training procedure improves the performance of ResMLP-S12 by $7.4\%$ compared to the training employed for ResNet~\cite{He2016ResNet}\footnote{Interestingly, if trained with this ``old-fashion'' setting, ResMLP-S12 outperforms AlexNet~\cite{Krizhevsky2012AlexNet} by a margin. }. This is in line with recent work pointing out the importance of the training strategy over the model choice~\cite{Bello2021RevisitingRI,Radosavovic2020RegNet}.
Pre-training on more data and distillation also improve the performance of ResMLP, especially for the bigger models, \eg, distillation improves the accuracy of ResMLP-B24/8 by $2.6\%$.

\textbf{Other analysis.} 
In our early exploration, we evaluated several alternative design choices. %
        As in transformers, we could use positional embeddings mixed with the input patches. 
        In our experiments   we did not see any benefit from using these features, see Appendix~\ref{sec:exploration}.
        This observation suggests that our cross-patch sublayer provides sufficient spatial communication, and 
        referencing absolute positions obviates the need for any form of positional encoding.

\subsection{Transfer learning} 
We evaluate the quality of features obtained from a \OURS architecture when transferring them to other domains. The goal is to assess if the features generated from a feedforward network are more prone to overfitting on the training data distribution.
We adopt the typical setting where we pre-train a model on \ImNet-1k and fine-tune it on the training set associated with a specific domain. %
We report the performance with different architectures on various image benchmarks in Table~\ref{tab:transfer}, namely CIFAR-10 and CIFAR-100~\cite{krizhevsky2009learning}, Flowers-102~\cite{Nilsback08}, Stanford Cars~\cite{Cars2013} and iNaturalist~\cite{Horn2019INaturalist}.  
We refer the reader to the corresponding references for a more detailed description of the  datasets.

We observe that the performance of our \OURS is competitive with the existing architectures, showing that pretraining feedforward models with enough data and regularization via data augmentation greatly reduces their tendency to overfit on the original distribution. Interestingly, this regularization also prevents them from overfitting on the training set of smaller dataset during the fine-tuning stage.
\def \mysp {\hspace{13pt}}
\begin{table}
    \hspace{-3pt}
    \scalebox{0.84}{
    \begin{tabular}{l@{\mysp}@{\mysp}c@{\mysp}@{\mysp}c@{\mysp}c@{\mysp}c@{\mysp}c@{\mysp}c@{\mysp}c@{\mysp}c}
    \toprule
    Architecture              
     & {FLOPs}
           & {Res.}
        & {CIFAR$_{10}$}
        & {CIFAR$_{100}$}  
        & {Flowers102} 
        & {Cars} 
        & {iNat$_{18}$} 
        & {iNat$_{19}$} 
       \\
    \midrule
    EfficientNet-B7~\cite{tan2019efficientnet} & \pzo37.0B & 600 & 98.9 & 91.7  & 98.8 & 94.7 & \_ & \_  \\
    ViT-B/16~\cite{dosovitskiy2020image}  & \pzo55.5B & 384 & 98.1 & 87.1 & 89.5 & \_   & \_   & \_   \\
    ViT-L/16~\cite{dosovitskiy2020image}  & 190.7B    & 384 & 97.9 & 86.4 & 89.7 & \_   & \_   & \_   \\
    Deit-B/16~\cite{Touvron2020TrainingDI} & \pzo17.5B & 224 & 99.1 & 90.8 & 98.4 & 92.1 & 73.2 & 77.7  \\
    ResNet50~\cite{Touvron2020GrafitLF} & \dzo4.1B & 224 & \_ & \_ & 96.2 & 90.0 & 68.4 & 73.7  \\
    Grafit/ResNet50~\cite{Touvron2020GrafitLF} & \dzo4.1B & 224 & \_ & \_ & 97.6 & 92.7 & 68.5 & 74.6  \\
    \midrule
    ResMLP-S12      & \dzo3.0B & 224 & 98.1 & 87.0 & 97.4 & 84.6 & 60.2 & 71.0 \\
    ResMLP-S24       & \dzo6.0B &224 & 98.7 & 89.5 & 97.9 & 89.5 & 64.3  & 72.5  \\
    \bottomrule
    \end{tabular}}
    \smallskip
    \caption{
    \textbf{Evaluation on transfer learning.}
    Classification accuracy (top-1) of models trained on \ImNet-1k  for transfer to datasets covering different domains.
    The \OURS architecture takes $224\!\times\!224$ images  during training and transfer, while  ViTs and EfficientNet-B7 work with higher resolutions, see ``Res.'' column. 
    }
    \label{tab:transfer}
\end{table}

\subsection{Machine translation}

We also evaluate the ResMLP transpose-mechanism to replace the self-attention in the encoder and decoder of a neural machine translation system.
We train models on the WMT 2014 English-German and English-French tasks, following the setup from Ott \etal \cite{ott2018scaling}.
We consider models of dimension 512, with a hidden MLP size of 2,048, and with 6 or 12 layers. Note that the current state of the art employs much larger models: our 6-layer model is more comparable to the base transformer model from Vaswani \etal \cite{vaswani2017attention}, which serves as a baseline, along with pre-transformer architectures such as recurrent and convolutional neural networks. 
We use Adagrad with learning rate 0.2, 32k steps of linear warmup, label smoothing 0.1, dropout rate 0.15 for En-De and 0.1 for En-Fr.
We initialize the LayerScale parameter to 0.2. We generate translations with the beam search algorithm, with a beam of size 4.
As shown in Table~\ref{tab:wmt}, the results are at least on par with the compared architectures. 

\begin{table}[h]
    \captionof{table}{\textbf{Machine translation} on WMT 2014 translation tasks. We report tokenized BLEU on \emph{newstest2014}.    \label{tab:wmt}}
    \def \mysp {\hspace{6pt}}
     \centering \scalebox{0.9}
    {
    \begin{tabular}{lcccccc}
    \toprule
    Models &  GNMT~\cite{wu2016google} & ConvS2S~\cite{gehring2017convolutional} & Transf. (base)~\cite{vaswani2017attention} &  ResMLP-6 & ResMLP-12 &    \\
    \midrule
    \textsc{En-De} & 24.6 & 25.2 &  27.3  & 26.4 & 26.8  &  \\
    \textsc{En-Fr} & 39.9 & 40.5 &  38.1  & 40.3 & 40.6  &  \\
    \bottomrule
    \end{tabular}}
\end{table}
\section{Related work }
\label{sec:related} 

\nocite{NIPS2015_6855456e}

We review the research on applying Fully Connected Network (FCN) for computer vision problems as well as other architectures that shares common modules with our model.

\textbf{Fully-connected network for images.}
Many studies have shown that FCNs are competitive with convnets for the tasks of digit recognition~\cite{cirecsan2012deep,simard2003best}, keyword spotting~\cite{chatelain2006extraction} and handwritting recognition~\cite{bluche2015deep}.
Several works~\cite{lin2015far,mocanu2018scalable,urban2016deep} have questioned if FCNs are also competitive on natural image datasets, such as CIFAR-10~\cite{krizhevsky2009learning}. 
More recently, d'Ascoli~\etal~\cite{d2019finding} have shown that a FCN initialized with the weights of a pretrained convnet achieves performance that are superior than the original convnet. 
Neyshabur~\cite{neyshabur2020towards} further extend this line of work by achieving competitive performance by training an FCN from scratch but with a regularizer that constrains the models to be close to a convnet. 
These studies have been conducted on small scale datasets with the purpose of studying the impact of architectures on  generalization in terms of sample complexity~\cite{du2018many} and energy landscape~\cite{keskar2016large}.
In our work, we show that, in the larger scale setting of ImageNet, FCNs can attain surprising accuracy without any constraint or initialization inspired by convnets. 

Finally, the application of FCN networks in computer vision have also emerged in the  study of the properties of networks with infinite width~\cite{novak2018bayesian}, or for inverse scattering problems~\cite{khoo2019switchnet}.
More interestingly, the Tensorizing Network~\cite{novikov2015tensorizing} is an approximation of very large FCN that shares similarity with our model, in that they intend to remove prior by approximating even more general tensor operations, \ie, not arbitrarily marginalized along some pre-defined sharing dimensions. 
However, their method is designed to compress the MLP layers of a standard convnets.

\textbf{Other architectures with similar components.}
Our FCN architecture shares several components with other architectures, such as convnets~\cite{Krizhevsky2012AlexNet,lecun1998gradient} or transformers~\cite{vaswani2017attention}.
A fully connected layer is equivalent to a convolution layer with a $1\times 1$ receptive field, and several work have explored convnet architectures with small receptive fields.
For instance, the VGG model~\cite{Simonyan2015VGG}  uses $3\!\times\! 3$ convolutions, and later, other architectures such as the ResNext~\cite{xie2017aggregated} or the Xception~\cite{chollet2017xception} mix $1\!\times\!1$ and $3\!\times\!3$ convolutions.
In contrast to convnets, in our model interaction between patches is obtained via a linear layer that is shared across channels, and that relies on absolute rather than relative positions. 

More recently, transformers have emerged as a promising  architecture for computer vision~\cite{ child2019generating, dosovitskiy2014discriminative,parmar2018image,Touvron2020TrainingDI,zhao2020exploring}.
In particular, our architecture takes inspiration from the structure used in the Vision Transformer~(ViT)~\cite{dosovitskiy2014discriminative}, and as consequence, shares many components.
Our model takes a set of non-overlapping patches 
as input and passes them through a series of MLP layers that share the same structure as ViT, replacing the self-attention layer with a linear patch interaction layer.
Both layers have a global field-of-view, unlike convolutional layers. Whereas in self-attention the weights to aggregate information from other patches are data dependent through queries and keys, in \OURS the weights are not data dependent and only  based on absolute positions of patches.  
In our implementation we follow the improvements of DeiT~\cite{Touvron2020TrainingDI} to train vision transformers, use the skip-connections from  ResNets~\cite{He2016ResNet} with pre-normalization of the layers~\cite{chen2018best,He2016IdentityMappings}.

Finally,  our work questions the importance of self-attention in %
existing architectures. %
Similar observations have been made in natural language processing.
Notably,  Synthesizer~\cite{tay2020synthesizer} shows that dot-product self-attention can be replaced by a feedforward network, with  competitive performance on sentence representation benchmarks.
As opposed to our work,  Synthesizer does use data dependent weights, but in contrast to transformers the weights are determined from the queries only.

\section{Conclusion}
\label{sec:conclusion}

In this paper we have shown that a simple residual architecture, 
whose residual blocks consist of a one-hidden layer feed-forward network and a linear patch interaction layer, 
achieves an unexpectedly high performance on ImageNet classification benchmarks, provided that we adopt a modern training strategy such as those recently introduced for transformer-based architectures. 
Thanks to their simple structure, with linear layers as the main mean of communication between patches, we can vizualize the filters  learned by this simple MLP. 
While some of the layers are similar  to convolutional filters, we also observe  sparse long-range interactions as early as the second layer of the network. 
We hope that our model free of spatial priors  will contribute to further  understanding of what  networks with less priors learn, and potentially guide the design choices of future networks without the  pyramidal design prior adopted by most convolutional neural networks.  

\section{Acknowledgments}

We would like to thank Mark Tygert for relevant  references. 
This work builds upon the Timm library~\cite{pytorchmodels} by Ross Wightman. 

\begingroup
    \small
    \bibliographystyle{ieee_fullname}
    \bibliography{egbib}
\endgroup

\clearpage

\appendix\newpage
\appendix

\clearpage
\counterwithin{figure}{section}
\counterwithin{table}{section}
\counterwithin{equation}{section}

\pagenumbering{Roman}  

\newpage
\vskip .375in
\begin{center}
{\Large \bf \inserttitle \\ \vspace{0.5cm} \large Appendix \par}
  \vspace*{24pt}
  {
  \par
  }
\end{center}

\section{Report on our exploration phase}
\label{sec:exploration}

As discussed in the main paper, our work on designing a residual multi-layer perceptron was inspired by the Vision Transformer. For our exploration, we have adopted the recent CaiT variant~\cite{touvron2021going} as a starting point. This transformer-based architecture achieves state-of performance with Imagenet-training only (achieving 86.5\% top-1 accuracy on Imagenet-val for the best model). Most importantly, the training is relatively stable with increasing depth.

In our exploration phase, our objective was to radically simplify this model. For this purpose, we have considered the Cait-S24 model for faster iterations. This network consists of 24-layer with a working dimension of 384. All our experiments below were carried out with images in resolution 224$\times$224 and $N=16\times 16$ patches. Trained with regular supervision, Cait-S24 attains 82.7\% top-1 acc. on Imagenet.  

\paragraph{SA $\rightarrow$ MLP. }
The self-attention can be seen a weight generator for a linear transformation on the values. 
Therefore, our first design modification was to get rid of the self-attention by replacing it by a residual feed-forward network, which takes as input the \emph{transposed} set of patches instead of the patches. In other terms, in this case we alternate residual blocks operating along the channel dimension with some operating along the patch dimension. In that case, the MLP replacing the self-attention consists of the sequence of operations 
\begin{center}
$(\cdot)^T$ --- linear $N\times4N$ --- GELU --- \text{linear} $4N\times N$ --- $(\cdot)^T$
\end{center} 

Hence this network is symmetrical in $N$ and $d$. By keeping the other elements identical to CaiT, the accuracy drops to $80.2\%$ (-2.5\%) when replacing self-attention layers. 

\paragraph{Class-attention $\rightarrow$ class-MLP. }
If we further replace the class-attention layer of CaiT by a MLP as described in our paper, then we obtain an attention-free network whose top-1 accuracy on Imagenet-val is 79.2\%, which is comparable to a ResNet-50 trained with a modern training strategy. 
This network has served as our baseline for subsequent ablations. Note that, at this stage, we still include LayerScale, a class embedding (in the class-MLP stage) and positional encodings. 

\paragraph{Distillation. }
The same model trained with distillation inspired by Touvron et al.~\cite{Touvron2020TrainingDI} achieves 81.5\%. The distillation variant we choose corresponds to the ``hard-distillation'', whose main advantage is that it does not require any parameter-tuning compared to vanilla cross-entropy. 
Note that, in all our experiments, this distillation method seems to bring a gain that is complementary and seemingly almost orthogonal to other modifications. 

\paragraph{Activation: LayerNorm $\rightarrow$ X. }
We have tried different activations on top of the aforementioned MLP-based baseline, and kept GeLU for its accuracy and to be consistent with the transformer choice. 
\begin{center}
\begin{tabular}{l|c}
\toprule
    Activation   & top-1 acc. \\
    \midrule
GeLU (baseline) & 79.2\% \\
SILU &  78.7\% \\
Hard Swish & 78.8\% \\
ReLU  &  79.1\% \\
\bottomrule
\end{tabular}
\end{center}

\paragraph{Ablation on the size of the communication MLP. } 

For the MLP that replaced the class-attention, we have explored different sizes of the latent layer, by adjusting the expansion factor $e$ in the sequence: linear $N\times e\times N$ --- GELU --- \text{linear} $e \times N\times N$. For this experiment we used average pooling to aggregating the patches before the classification layer. 
\begin{center}
\begin{tabular}{l|cccccc}
\toprule
expansion factor $\times e$  & $\times$0.25 & $\times$0.5  & $\times$1  & $\times$2  & $\times$3  & $\times$4 \\
Imnet-val top-1 acc. & 78.6 & 79.2 & 79.2 & 79.3 & 78.8 & 78.8 \\
\bottomrule
\end{tabular}
\end{center}
We observe that a large expansion factor is detrimental in the patch communication, possibly because we should not introduce too much capacity in this residual block. This has motivated the choice of adopting a simple linear layer of size $N\times N$: This subsequently improved performance to $79.5\%$ in a setting  comparable to the table above. 
Additionally, as shown earlier this choice allows visualizations of the interaction between patches. 

\paragraph{Normalization. } 
On top of our MLP baseline, we have tested different variations for  normalization layers. We report the variation in performance below. 

\begin{center}
\begin{tabular}{l|c}
\toprule
    Pre-normalization      & top-1 acc. \\
    \midrule
    Layernorm (baseline)   & 79.2\%  \\
    Batch-Norm             & \ +0.8\% \\
    $\ell_2$-norm          & \ +0.4\% \\
    no norm (\texttt{Aff}) & \ +0.4\% \\
\bottomrule
\end{tabular}
\end{center}

For the sake of simplicity, we therefore adopted only the \texttt{Aff} transformation so as to not depend on any batch or channel statistics.

\paragraph{Position encoding.} 
In our experiments, removing the position encoding does not change the results when using a MLP or a simple linear layer as a communication mean across patch embeddings. This is not surprising considering that the linear layer implicitly encodes each patch identity as one of the dimension, and that additionally the linear includes a bias that makes it possible to differentiate the patch positions before the shared linear layer. 

\section{Analysis of interaction layers in 12-layer networks}
\label{sec:vizu_unsup}

In this section we further analyze the linear interaction layers in 12-layer models.

\begin{figure}[p]
\newcommand{\myvisu}[1]{{ \includegraphics[width=.3\linewidth]{figs/12layers_sup/layer_#1_w_fc1.pdf}}}
\small
\begin{tabular}{ccc}
       \myvisu{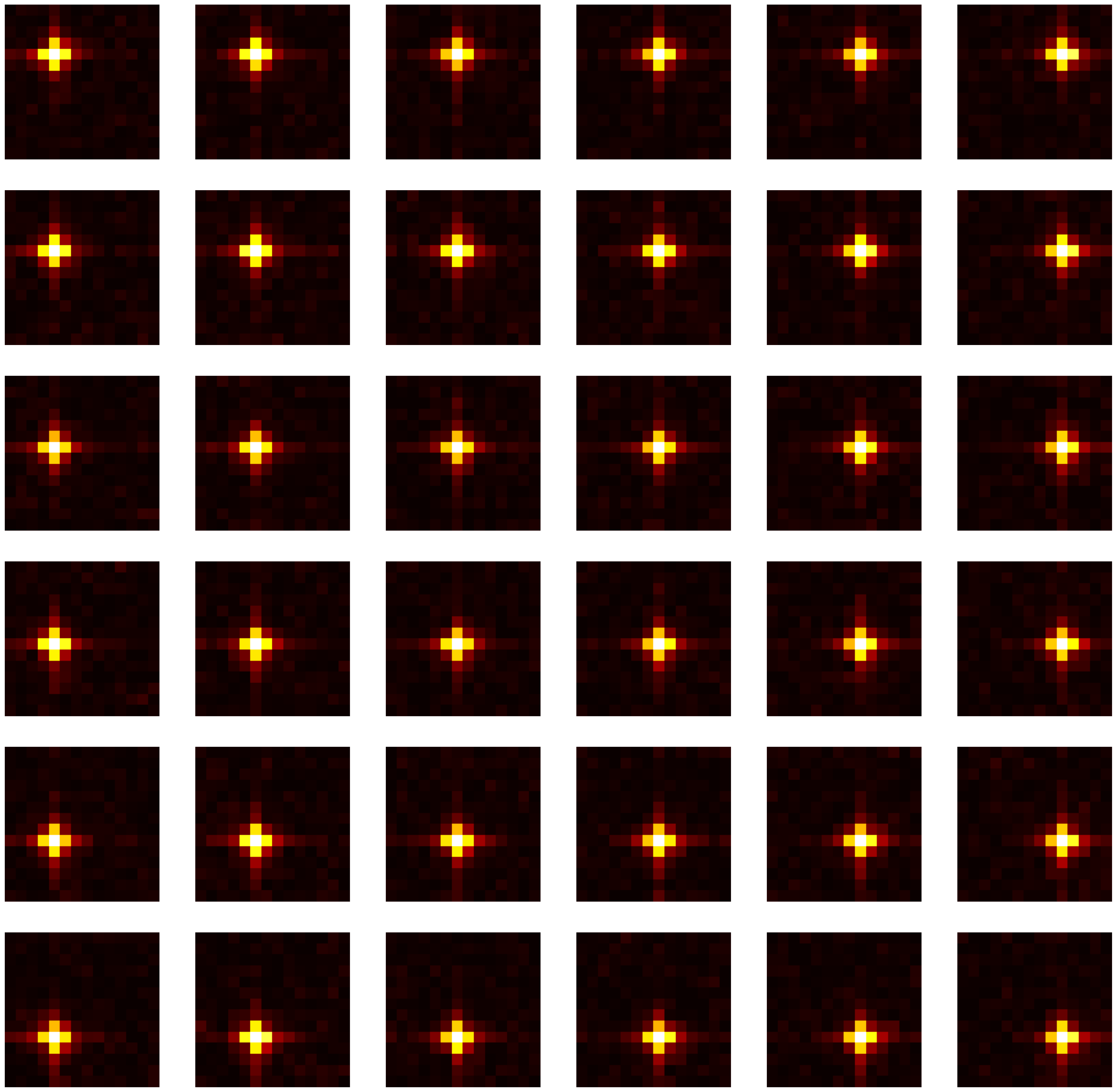} & \myvisu{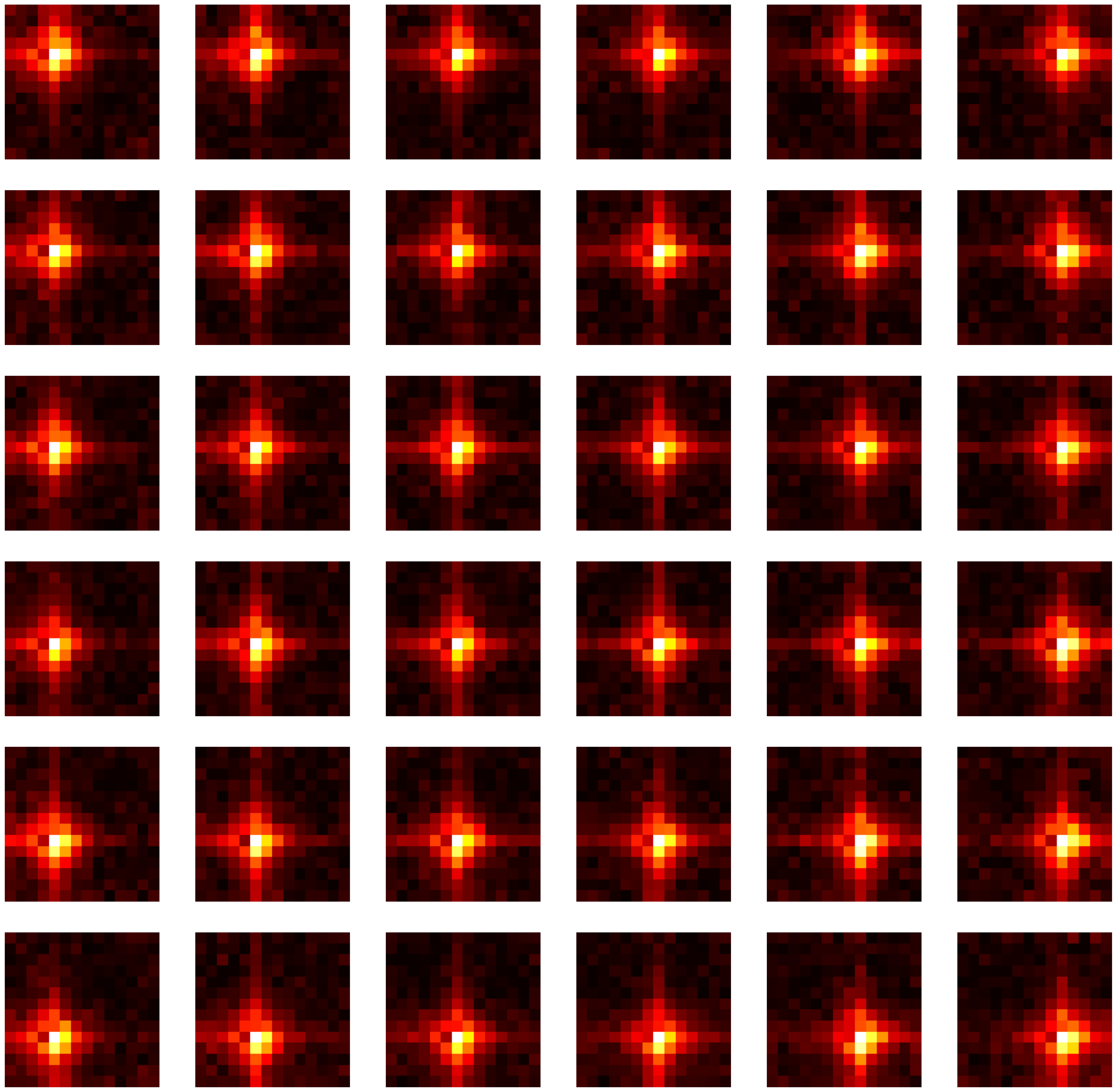} & \myvisu{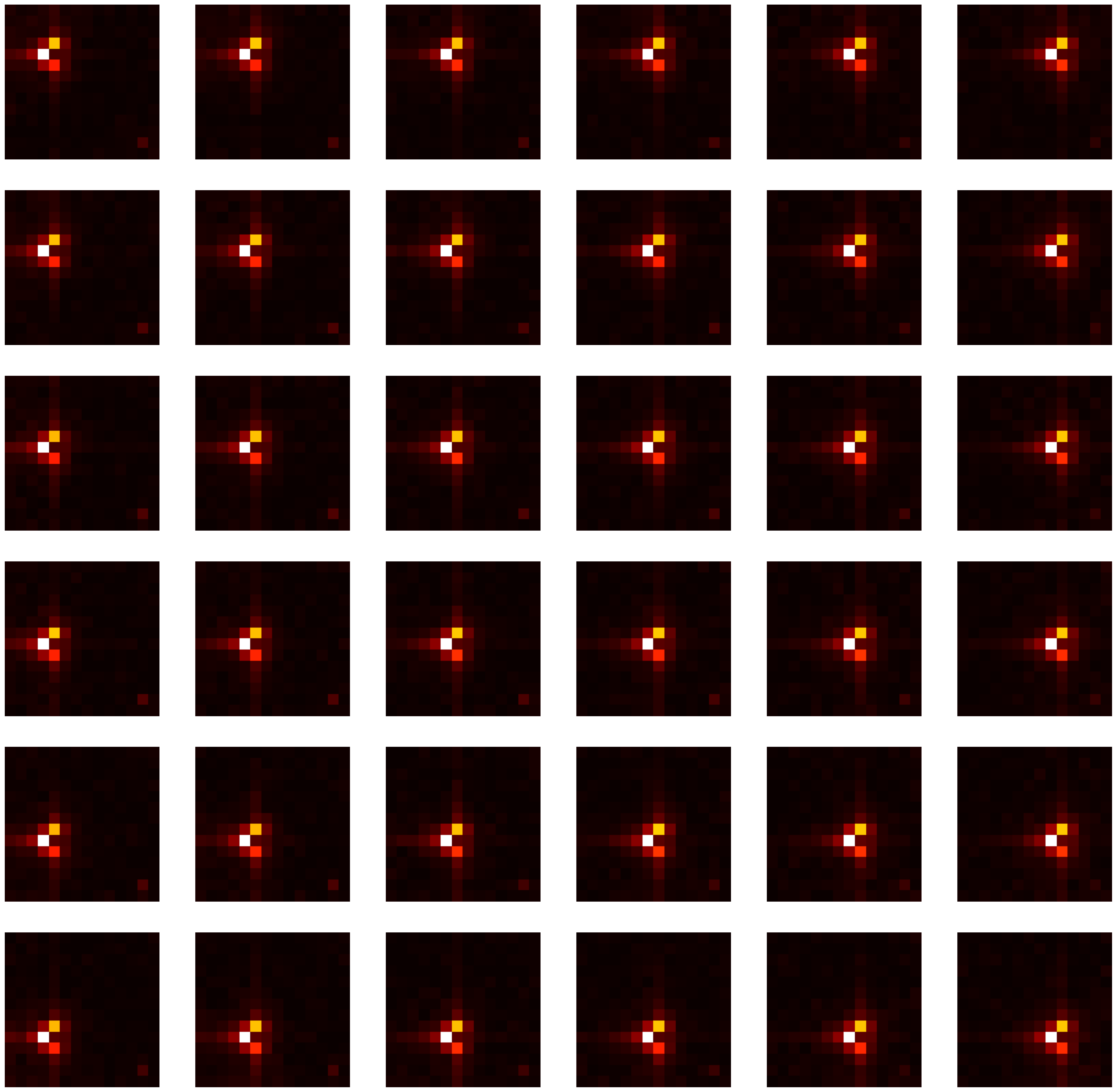}\\
       Layer 1 & Layer 2 & Layer 3 \\
  \myvisu{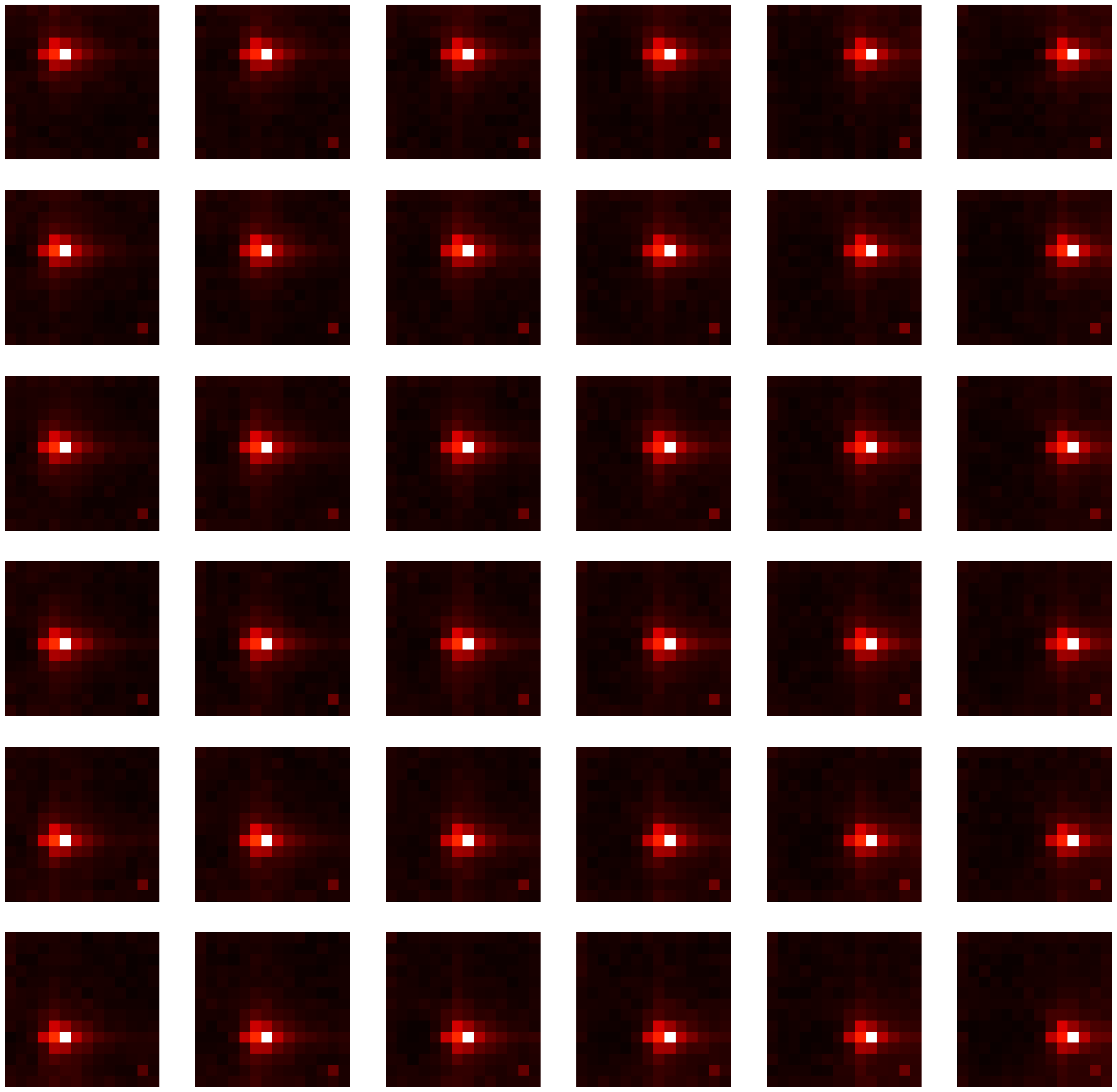} & \myvisu{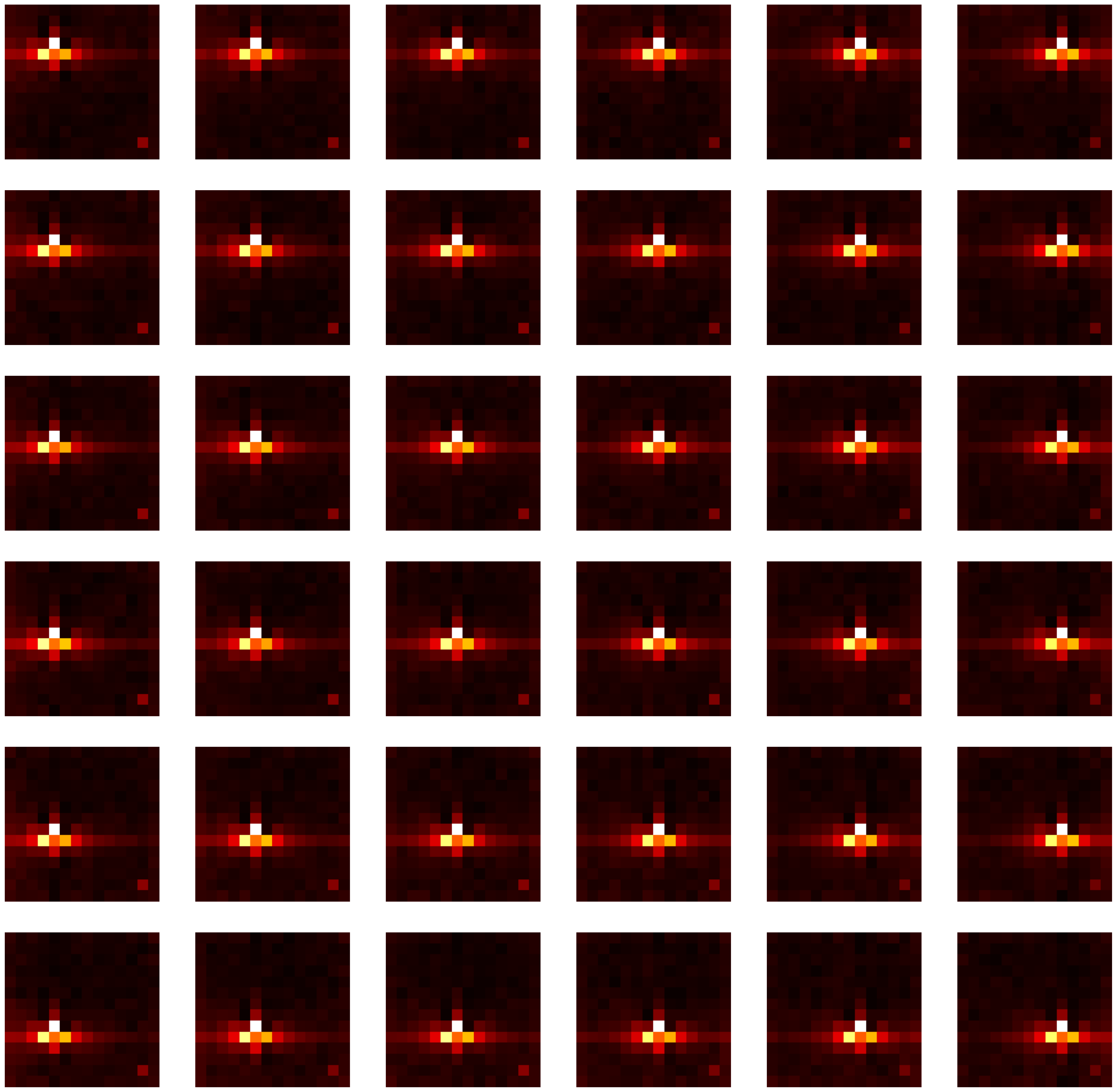} & \myvisu{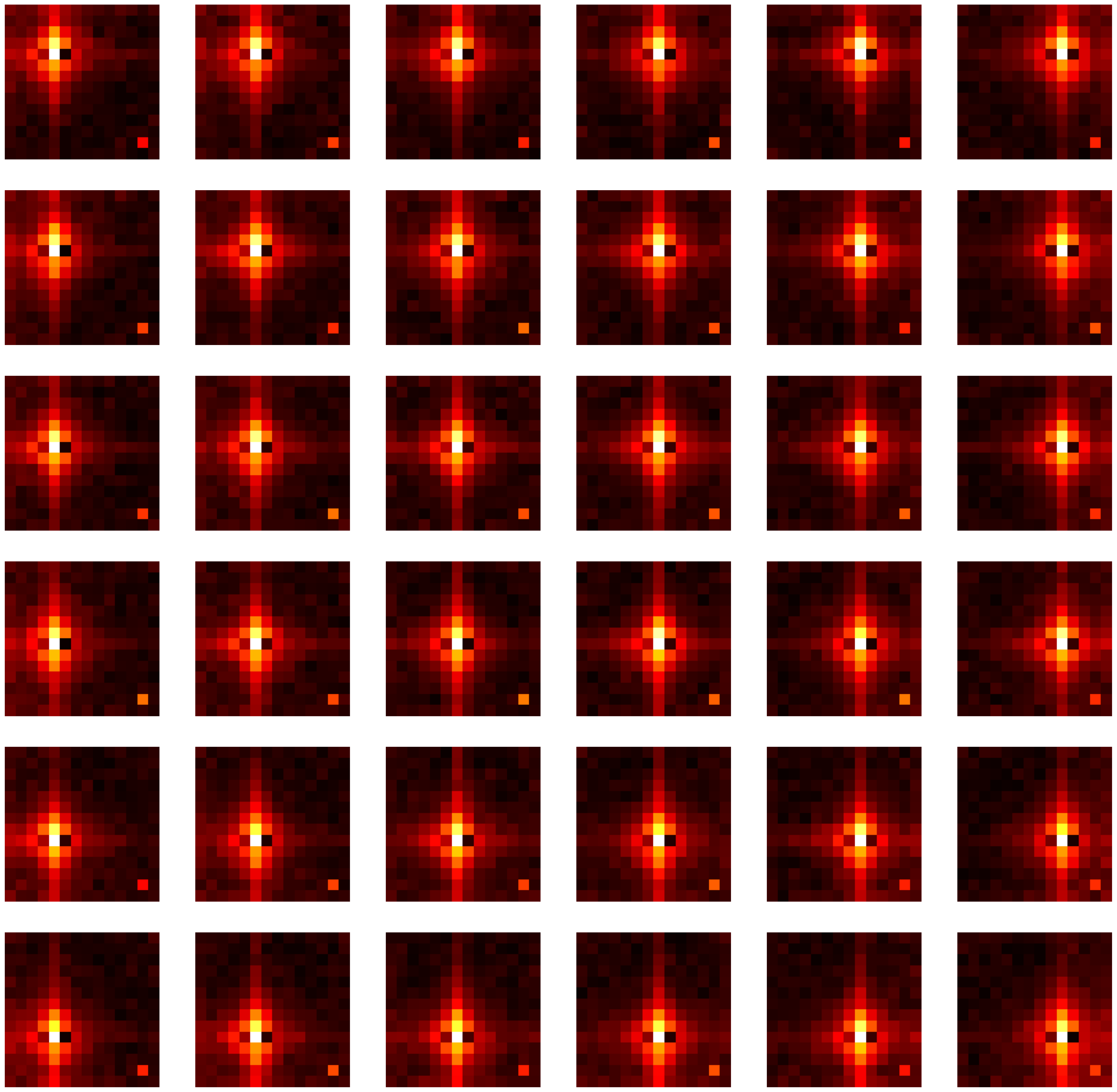}\\
  Layer 4 & Layer 5 & Layer 6 \\ 
       \myvisu{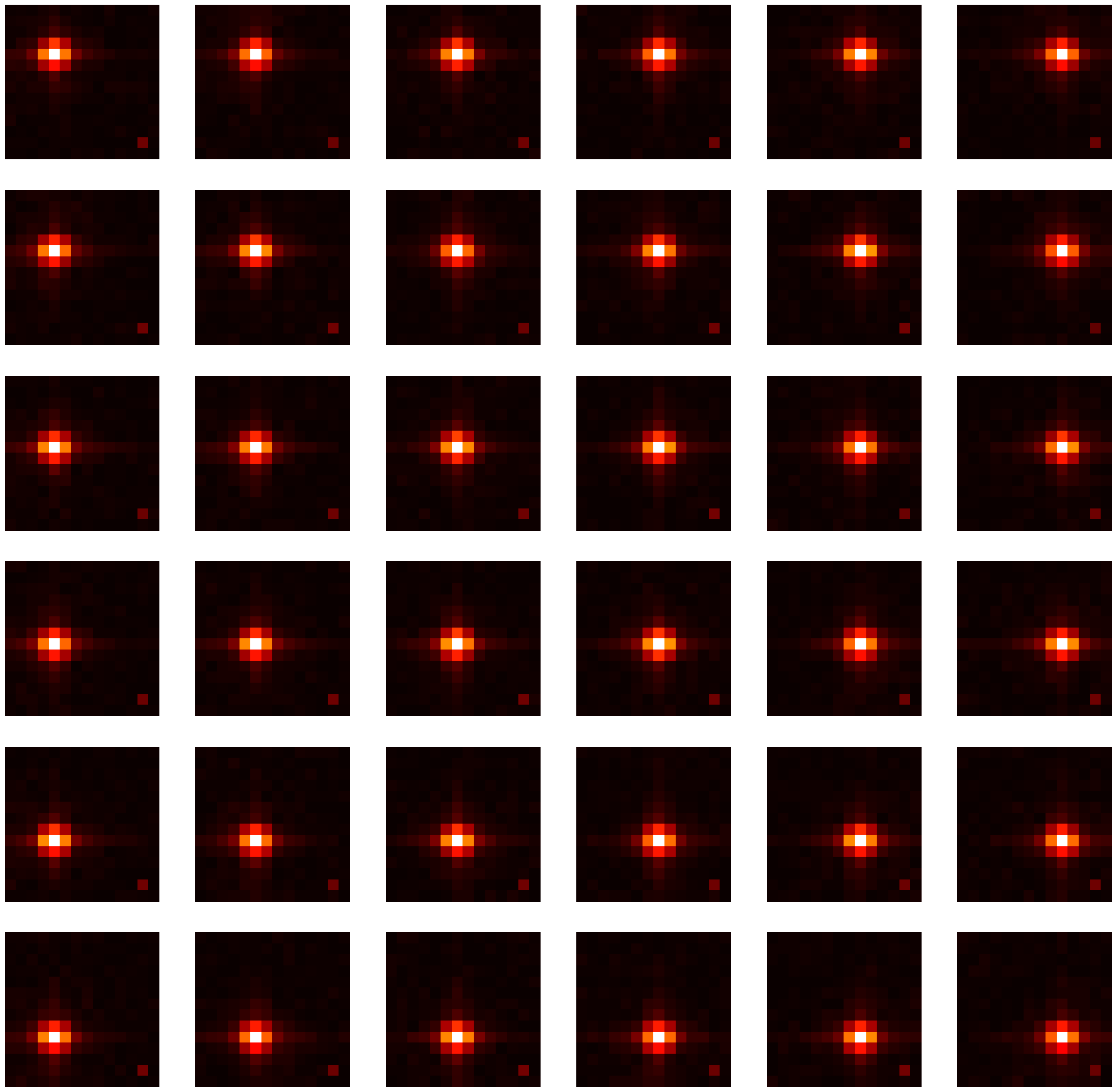} & \myvisu{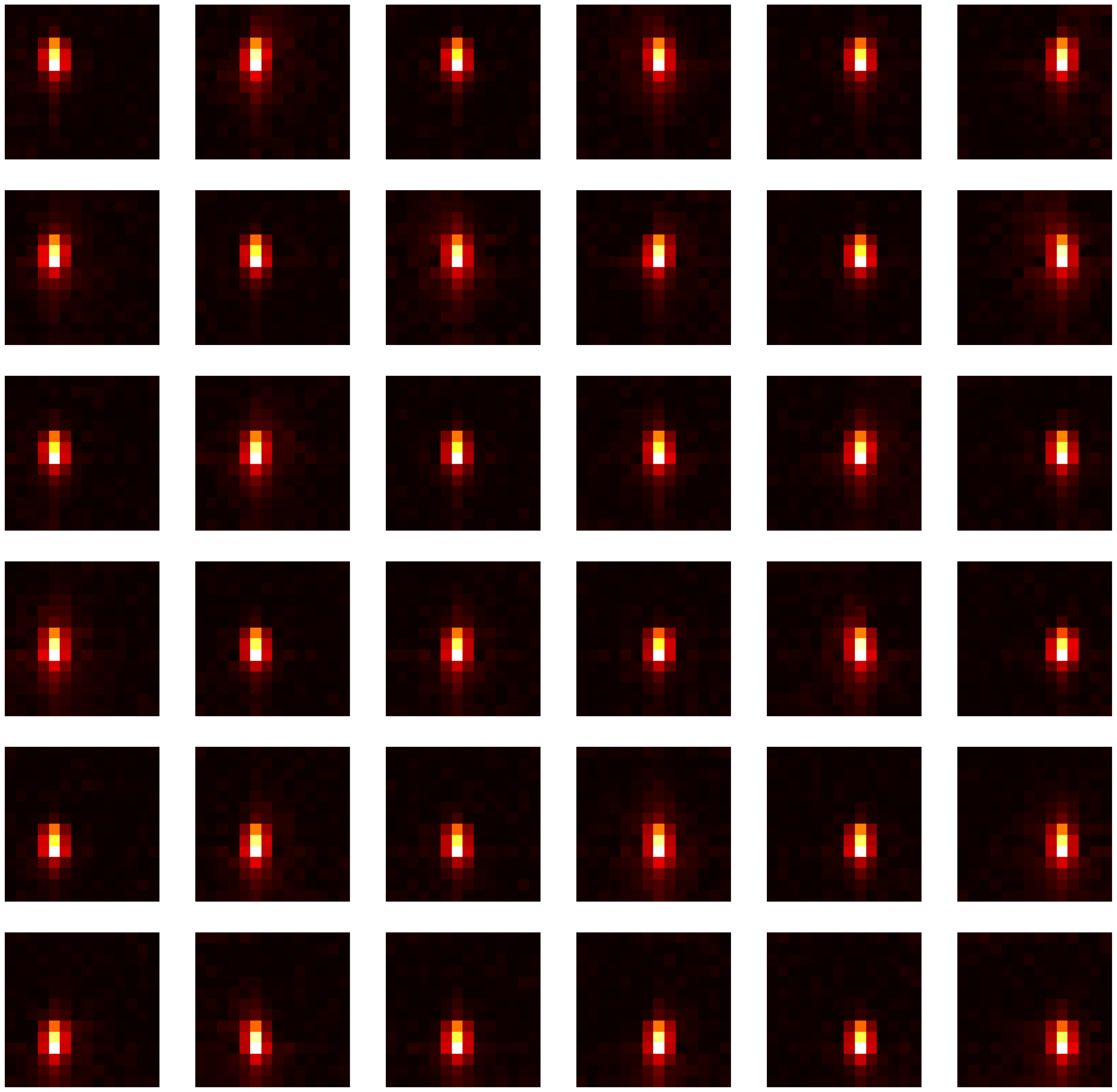} & \myvisu{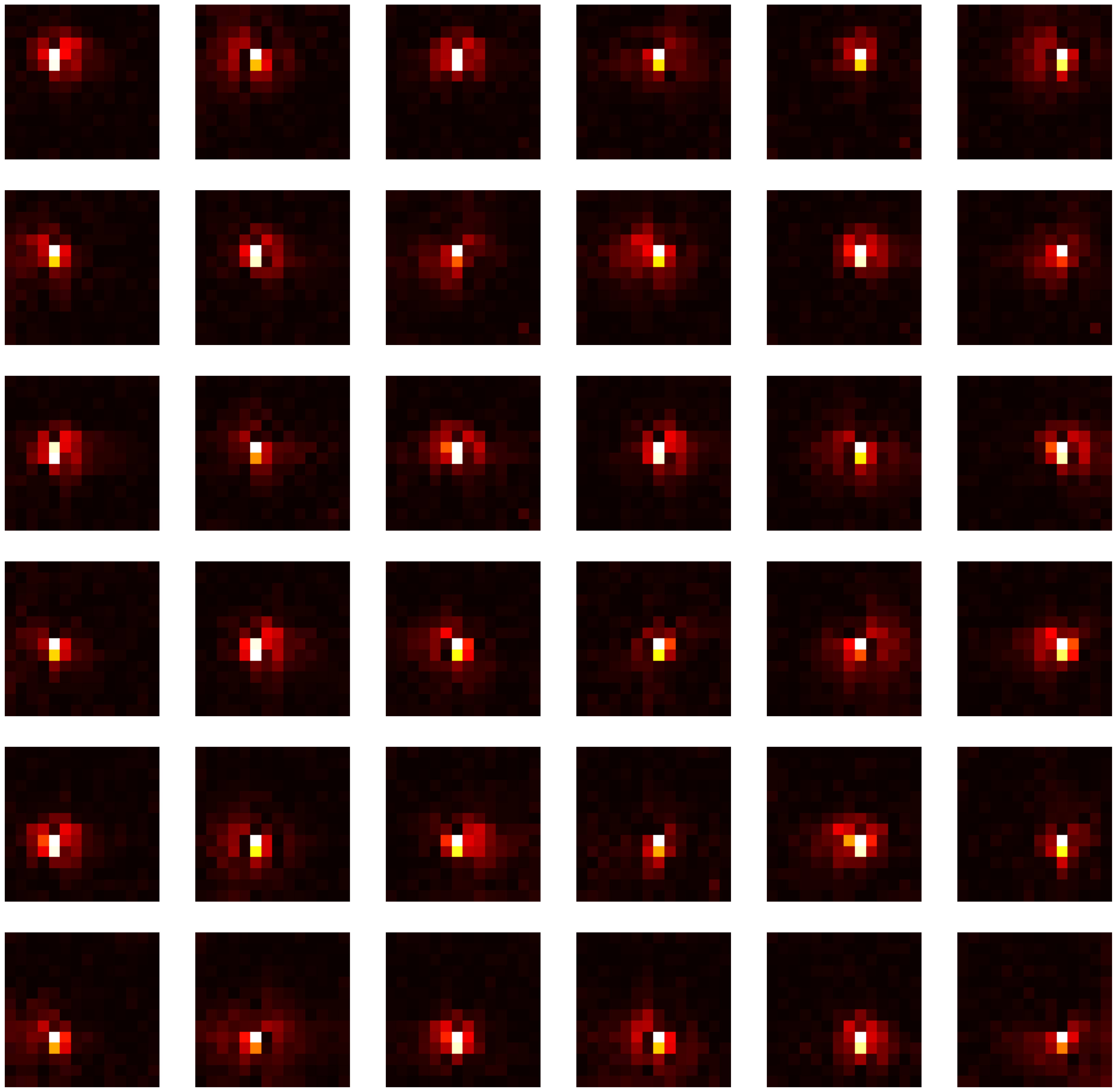}\\
       Layer 7 & Layer 8 & Layer 9 \\
  \myvisu{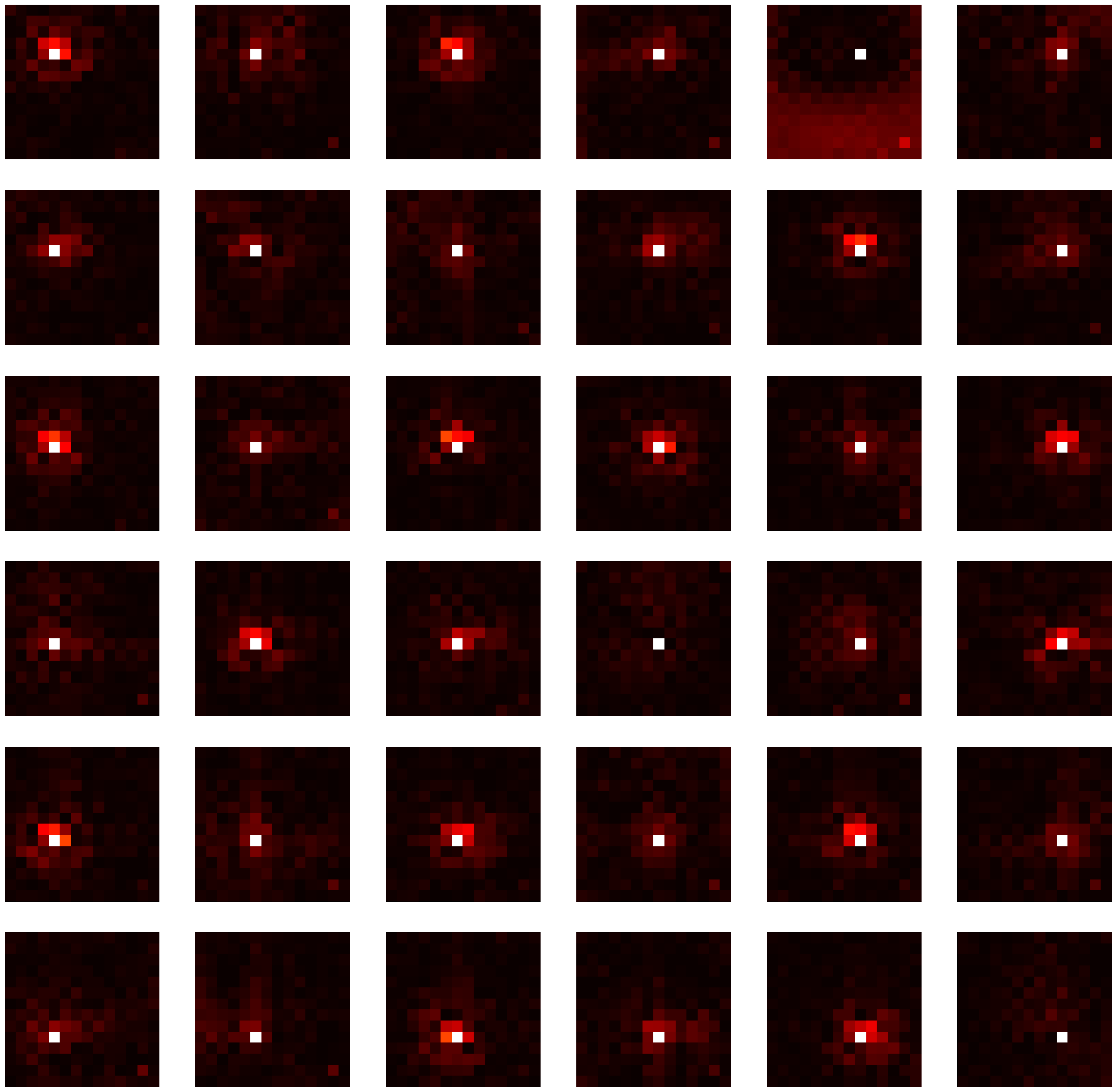} & \myvisu{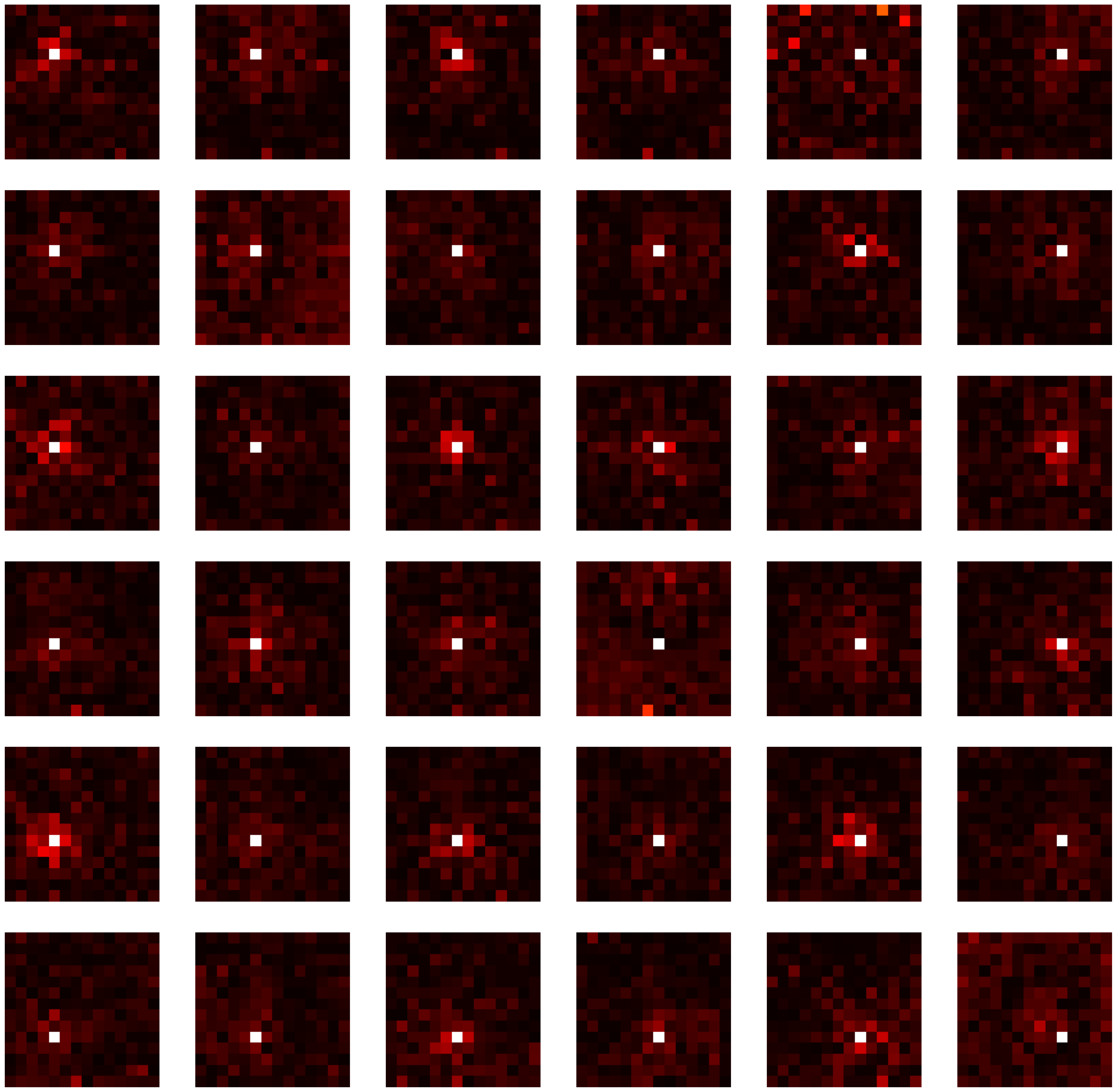} & \myvisu{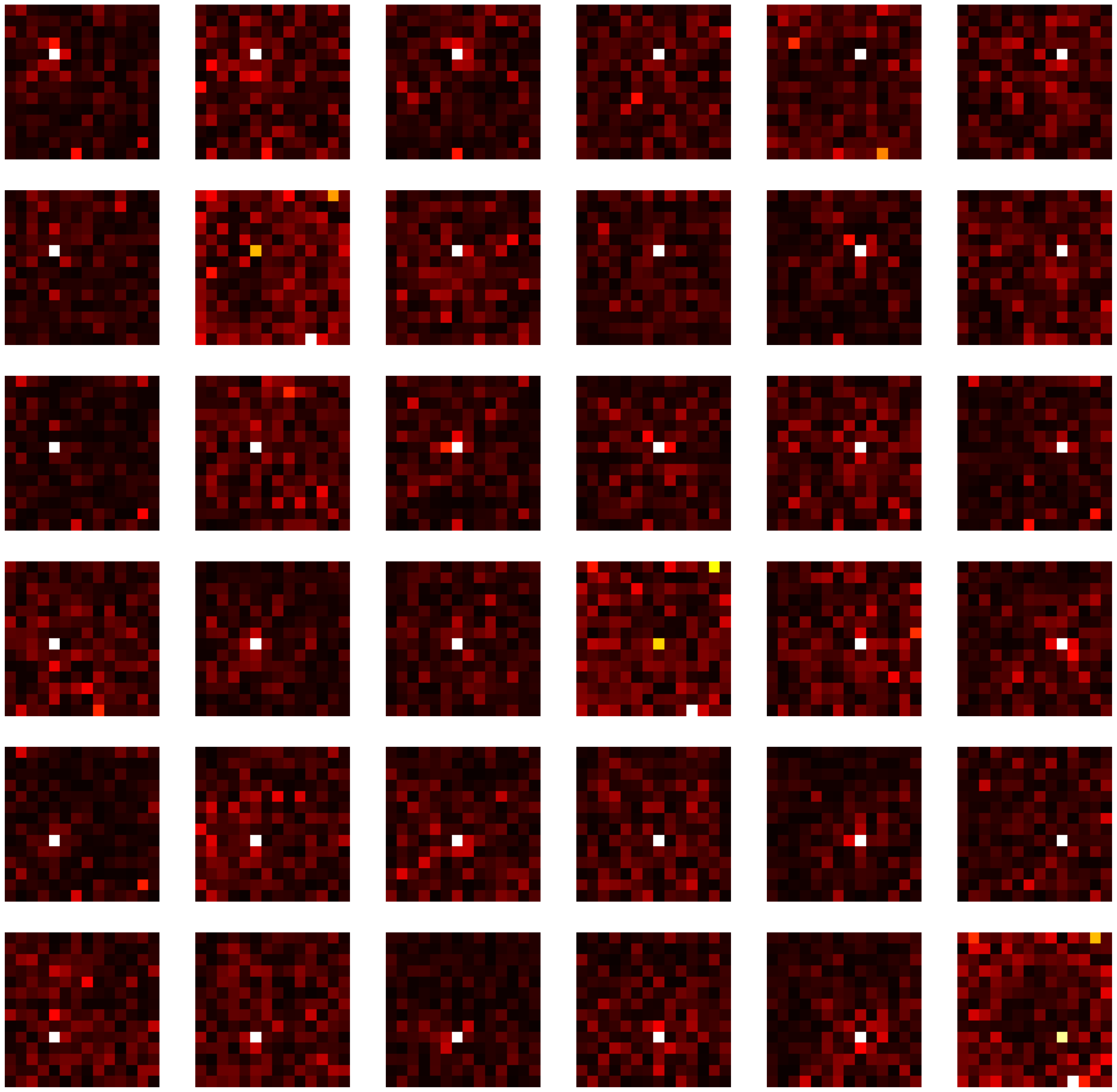}\\
  Layer 10 & Layer 11 & Layer 12 \\ 
\end{tabular}
 \caption{
{\bf Visualisation of the linear interaction layers in the supervised \OURS-S12 model.}  
For each layer we visualise the rows of the matrix $\mathbf{A}$ as a set of $14\times 14$ pixel images, for sake of space we only show the rows corresponding to the 6$\times$6 central patches. 
\label{fig:vizu_sup12}}
\end{figure}

In Figure \ref{fig:vizu_sup12} we  consider a ResMLP-S12 model trained on the \ImNet-1k dataset, as explained in Section~\ref{sec:experimental_setting}, and show all the 12 linear patch interaction layers. 
The linear interaction layers in the supervised 12-layer model are similar to those observed in the 24-layer model in  Figure \ref{fig:vizu}.

\begin{figure}[h]
\begin{minipage}{0.5\linewidth}
    \includegraphics[height=6.7cm]{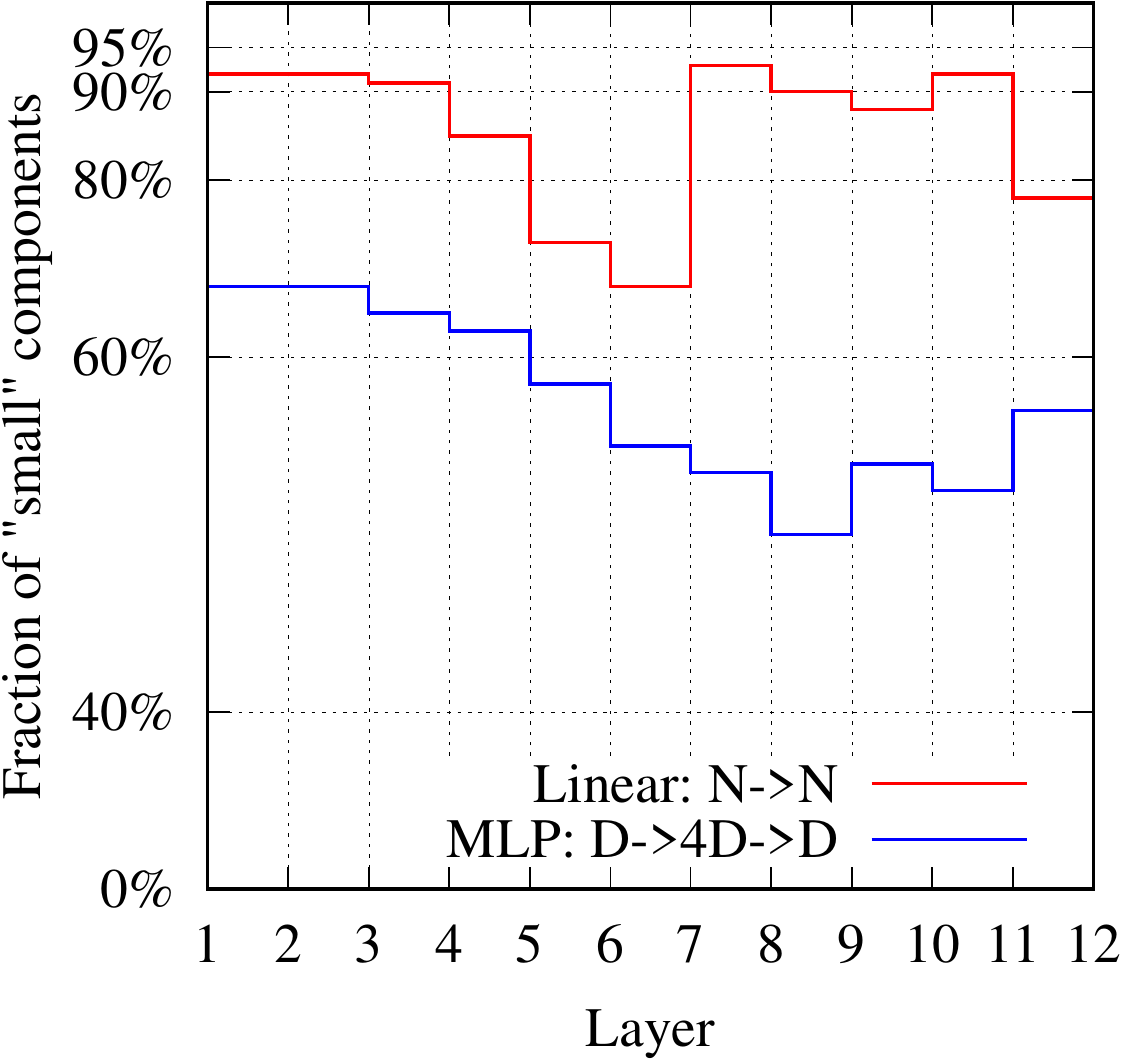} \\
    \centering \small Supervised
\end{minipage}
\hfill
\begin{minipage}{0.45\linewidth}
    \includegraphics[height=6.7cm,trim=50 0 0 0,clip]{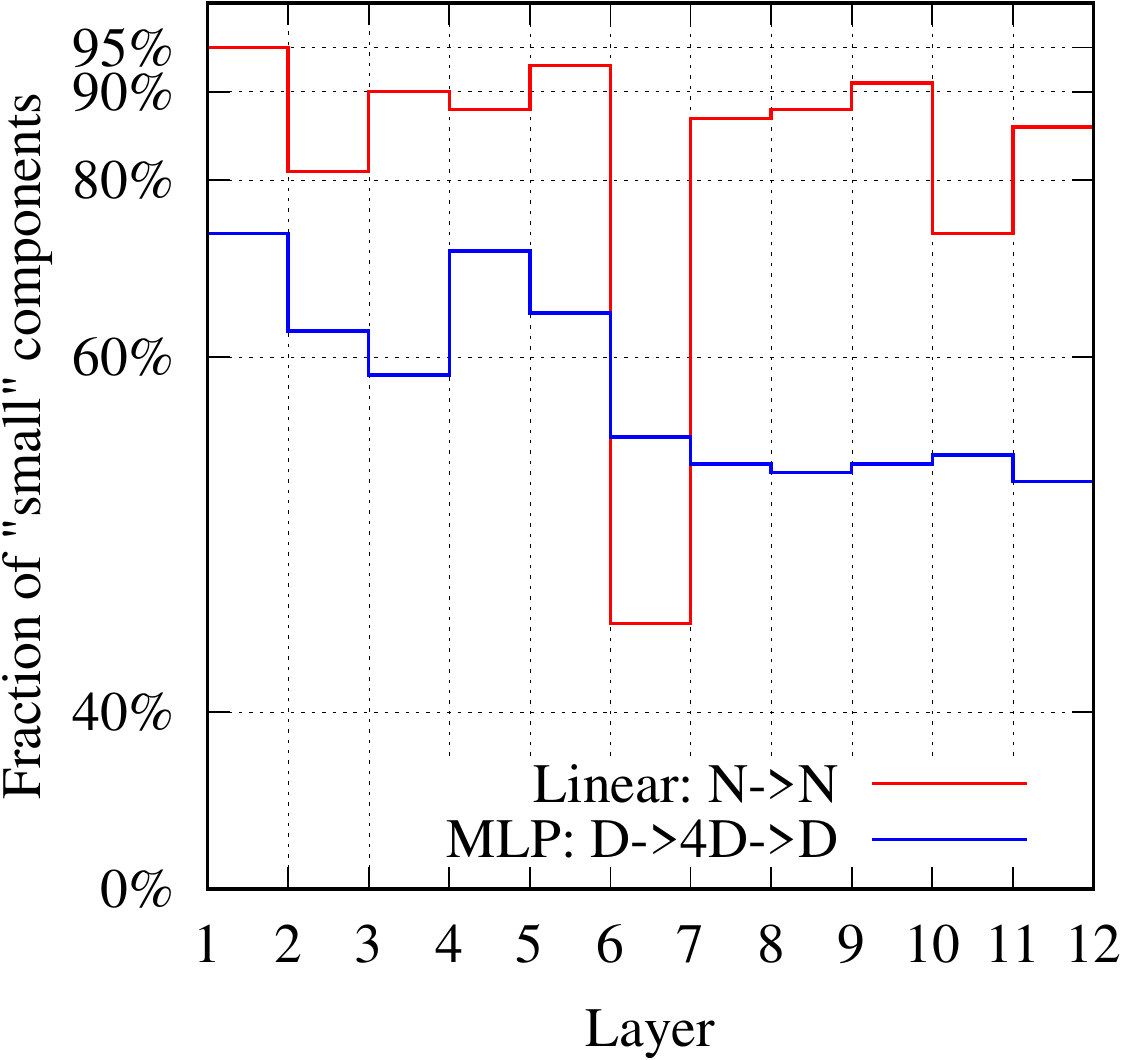} \\
    \centering \small  Distilled from convnet 
\end{minipage}
\caption{Degree of sparsity (fraction of values small than 5\% of the maximum) for \textcolor{red!80}{Linear} and \textcolor{blue!80}{MLP} layers, for  \OURS-S12 networks. The network trained in supervised mode  and the one learned with distillation overall have  a comparable degree of sparsity. The self-supervised model, trained during 300 epochs \vs 400 for the other ones, is  less sparse on the patch communication linear layer. 
 \label{fig:sparsity12}}    
\end{figure}

We also provide the corresponding sparsity measurements for  this model in Figure~\ref{fig:sparsity12}, analogous to the measurements in  Figure~\ref{fig:sparsity} for the supervised 24-layer model. 
The sparsity levels in the supervised 12-layer model (left panel) are similar to those observes in the supervised 24-layer model, cf.\ Figure~\ref{fig:sparsity}.
In the right panel of Figure~\ref{fig:sparsity12} we  consider the sparsity levels of the Distilled 12-layer model, which are overall similar to those observed for supervised the 12-layer and 24-layer models.

\clearpage 

\section{Model definition in Pytorch}
\label{sec:modelcode}

In Algorithm~\ref{alg:pseudocode} we provide the pseudo-pytorch-code associated with our model. 

\begin{algorithm}[h!]
\caption{Pseudocode of ResMLP in PyTorch-like style}
\label{alg:code}
\definecolor{codeblue}{rgb}{0.25,0.5,0.5}
\lstset{
  backgroundcolor=\color{white},
  basicstyle=\fontsize{7.2pt}{7.2pt}\ttfamily\selectfont,
  columns=fullflexible,
  breaklines=true,
  captionpos=b,
  commentstyle=\fontsize{7.2pt}{7.2pt}\color{codeblue},
  keywordstyle=\fontsize{7.2pt}{7.2pt},
}
\begin{lstlisting}[language=python]
# No norm layer
class Affine(nn.Module): 
    def __init__(self, dim):
        super().__init__()
        self.alpha = nn.Parameter(torch.ones(dim))
        self.beta  = nn.Parameter(torch.zeros(dim))
    def forward(self, x):
        return self.alpha * x + self.beta

# MLP on channels
class Mlp(nn.Module):
    def __init__(self, dim):
        super().__init__()
        self.fc1 = nn.Linear(dim, 4 * dim)
        self.act = nn.GELU()
        self.fc2 = nn.Linear(4 * dim, dim)
    def forward(self, x):
        x = self.fc1(x)
        x = self.act(x)
        x = self.fc2(x)
        return x
        
# ResMLP blocks: a linear between patches + a MLP to process them independently
class ResMLP_BLocks(nn.Module):
    def __init__(self, nb_patches ,dim, layerscale_init):
        super().__init__()
        self.affine_1 = Affine(dim)
        self.affine_2 = Affine(dim)
        self.linear_patches = nn.Linear(nb_patches, nb_patches) #Linear layer on patches
        self.mlp_channels = Mlp(dim) #MLP on channels
        self.layerscale_1 = nn.Parameter(layerscale_init * torch.ones((dim))) #LayerScale 
        self.layerscale_2 = nn.Parameter(layerscale_init * torch.ones((dim))) # parameters

    def forward(self, x):
        res_1 =  self.linear_patches(self.affine_1(x).transpose(1,2)).transpose(1,2)
        x = x + self.layerscale_1 * res_1
        res_2 = self.mlp_channels(self.affine_2(x))
        x = x + self.layerscale_2 * res_2
        return x   
        
# ResMLP model: Stacking the full network       
class ResMLP_models(nn.Module):
    def __init__(self, dim, depth, nb_patches, layerscale_init, num_classes):
        super().__init__()
        self.patch_projector = Patch_projector()
        self.blocks = nn.ModuleList([
            ResMLP_BLocks(nb_patches ,dim, layerscale_init)
            for i in range(depth)])
        self.affine = Affine(dim)
        self.linear_classifier = nn.Linear(dim, num_classes) 

    def forward(self, x):
        B, C, H, W = x.shape
        x = self.patch_projector(x) 
        for blk in self.blocks:
            x  = blk(x)
        x = self.affine(x)
        x = x.mean(dim=1).reshape(B,-1) #average pooling
        return self.linear_classifier(x)
\end{lstlisting}
\label{alg:pseudocode}
\end{algorithm}

\section{Additional Ablations}

\paragraph{Training recipe.}
DeiT~\cite{Touvron2020TrainingDI} proposes a  training strategy which allows for data-efficient vision transformers on ImageNet only.
In Table~\ref{tab:ablations_training} we ablate each component of the DeiT training to go back to the initial ResNet50 training.
As to be expected, the training used in the ResNet-50 paper~\cite{He2016ResNet} degrades the performance. 
\def \myminus {\textcolor{red!80}{\scalebox{0.7}{\fbox{\small \bf $-$}}\ \xspace}}
\def \myplus {\textcolor{blue!80}{\scalebox{0.7}{\fbox{\small \bf $+$}}\ \xspace}}

\begin{table}
    \centering
        \scalebox{0.9}{
        \begin{tabular}{ l  c }
             \toprule
              \multirow{2}{*}{Ablation}  & Imagenet1k-val   \\
                 & top1-acc (\%) \\
             \midrule
               DeiT-style training      & 76.6 \\
               \midrule
               \myminus CutMix~\cite{Yun2019CutMix} - Mixup~\cite{Zhang2017Mixup} & 76.0\\
               \myminus Random-erasing~\cite{Zhong2020RandomED} & 75.9\\
               \myminus RandAugment~\cite{Cubuk2019RandAugmentPA} & 73.2 \\
               SGD optimizer & 72.1\\
               \myminus Stochastic-depth~\cite{Huang2016DeepNW} & 70.7\\
               \myminus Repeated-augmentation~\cite{hoffer2020augment} & 69.4 \\
               120 epochs & 67.7 \\
               Step decay & 63.5 \\
               Batch size 256 & 69.3 \\
               \midrule
               ResNet-50 training 90 epochs & 69.2 \\
             \bottomrule
        \end{tabular}
        }
    \caption{Ablations on the training strategy }
    \label{tab:ablations_training}
\end{table}

\paragraph{Training schedule.}
Table~\ref{tab:ablation_epochs} compares the performance of ResMLP-S36 according to the number of training epochs.
We observe a saturation of the performance after 800 epochs for ResMLP. 
This saturation is observed in DeiT from 400 epochs. 
So ResMLP needs more epochs to be optimal.
\begin{table}[]
    \small \centering
    \scalebox{0.99}{
    \begin{tabular}{c|ccccc}
    \toprule
    Epochs & 300 & 400 & 500 & 800 & 1000 \\
    \midrule
    Inet-val  & 79.3 & 79.7 & 80.1  & \textbf{80.4} & 80.3 \\
    Inet-real & 85.5 & 85.6 & \textbf{85.9}  & 85.8 & 85.7 \\
    Inet-V2   & 68.0 & 68.4 &  68.4 & 68.9 & \textbf{69.0}\\
    \bottomrule
    \end{tabular}}
    \smallskip
    \caption{We compare the performance of ResMLP-S36 according to the number of training epochs. 
    \label{tab:ablation_epochs}}
\end{table}

\paragraph{Pooling layers.}

Table~\ref{tab:ablation_pooling} compares the performance of two pooling layers: average-pooling and class-MLP, with different depth with and without distillation.
We can see that class-MLP performs much better than average pooling by changing only a few FLOPs and number of parameters.
Nevertheless, the gap seems to decrease between the two approaches with deeper models.

\begin{table}[]
    \small \centering
    \scalebox{0.99}{
    \begin{tabular}{c|cc|cc|ccc}
    \toprule
    \#layers & Params & Flops & Training  & Pooling  & \multicolumn{3}{c}{top-1 acc. on \ImNet} \\
     & $\times 10^6$ & $\times 10^9$ &    & layer & Inet-val & Inet-real & Inet-V2 \\
    \midrule
    12  & 15.4 & 3.0 &  regular       & average  &  76.6 &  83.3 & 64.4 \\ 
    24 & 30.0 & 6.0 &  regular        & average   & 79.4 & 85.3 & 67.9  \\ 
    36 & 44.7 & 8.9 &  regular       & average    & 79.7 & 85.6 & 68.4 \\ 
    \midrule
    12 & 17.7 & 3.0 &   regular  & \tblue{class-MLP}  & 77.5 & 84.0 & 66.1 \\
    24 & 32.4 & 6.0 &   regular  & \tblue{class-MLP}  & 79.8 & 85.6 & 68.6  \\
    36 & 47.1 & 8.9 &   regular  & \tblue{class-MLP}  & 80.5 & 86.3 & 69.5  \\    
\midrule
\midrule
    12 &  15.4 & 3.0 & \tblue{distillation}  & average   & 77.8 & 84.6 & 66.0 \\ 
    24 & 30.0 & 6.0 &  \tblue{distillation}  & average   &  80.8 & 86.6 & 69.8 \\ 
    36 & 44.7 & 8.9 &  \tblue{distillation}  & average   & 81.0 & 86.8 & 70.2 \\ 
    \midrule
    12  &  17.7 & 3.0  &   \tblue{distillation}     & \tblue{class-MLP}  & 78.6 & 85.2 & 67.3  \\
    24  &  32.4 & 6.0  &   \tblue{distillation}     & \tblue{class-MLP}  & 81.1 & 87.0 & 70.4  \\
    36  &  47.1 & 8.9  &   \tblue{distillation}      & \tblue{class-MLP}  & 81.4 & 87.3  & 70.7 \\
    \bottomrule
    \end{tabular}}
    \smallskip
    \caption{We compare the performance of two pooling layers: average-pooling and class-MLP, with ResMLP-S architecture. We compare  different depth, regular training and distillation. 
    \label{tab:ablation_pooling}}
\end{table}

\end{document}